\documentclass{article}

\usepackage[main, final]{neurips_2026}

\usepackage[utf8]{inputenc} %
\usepackage[T1]{fontenc}    %
\usepackage{hyperref}       %
\usepackage{url}            %
\usepackage{booktabs}       %
\usepackage{amsfonts}       %
\usepackage{nicefrac}       %
\usepackage{microtype}      %
\usepackage{xcolor}         %
\usepackage{graphicx}

\usepackage{enumitem}

\usepackage{amsmath}
\usepackage{amssymb}
\usepackage{mathtools}
\usepackage{amsthm}
\usepackage{tikz}
\usetikzlibrary{shapes, arrows, positioning}
\usetikzlibrary{calc}

\usepackage{subcaption}
\usepackage{caption}
\usepackage{bm}

\usetikzlibrary{bayesnet}

\usepackage{bigdelim} 
\usepackage{wrapfig}

\usepackage[dvipsnames]{xcolor}

\definecolor{CBIndigo}{HTML}{332288}
\definecolor{CBGreen}{HTML}{117733}
\definecolor{CBPurple}{HTML}{AA4499}
\definecolor{CBWine}{HTML}{882255}

\usepackage[capitalize,noabbrev]{cleveref}

\theoremstyle{plain}
\newtheorem{theorem}{Theorem}[section]

\newtheorem{lemma}[theorem]{Lemma}
\newtheorem{corollary}[theorem]{Corollary}
\theoremstyle{definition}
\newtheorem{definition}[theorem]{Definition}

\theoremstyle{remark}

\title{2-Step Agent: How a Bayesian Decision Maker Learns from AI-Decision Support}

\author{
Otto Nyberg\thanks{Equal contribution}\hspace{0.4em}$^{1, 2}$
\And
Davide Tugnoli$^{* 3}$
\And
Fausto Carcassi$^{* 4}$
\And
Giovanni Cinà$^{* 1, 2, 4}$ \\[1ex]
$^1$ Department of Medical Informatics, Amsterdam UMC, University of Amsterdam, the Netherlands\\
$^2$ Amsterdam Public Health Research Institute, the Netherlands\\
$^3$ Department of Mathematics, Informatics and Geosciences, University of Trieste, Italy\\
$^4$ Institute for Logic, Language and Computation, University of Amsterdam, the Netherlands\\
Correspondence to: Otto Nyberg
<o.e.nyberg@amsterdamumc.nl>
}

\begin{document}

\maketitle

\setcounter{footnote}{0}

\begin{abstract}
Prognostic predictions from ML models support human decision making in several fields, including high-stakes ones such as healthcare and the judiciary.
Yet, we still lack a clear understanding of how decision makers learn from ML-based decision support (ML-DS).
In this paper, we introduce a general computational framework, \textit{the 2-Step Agent}, to capture how a prediction from an ML model can also be used as a source of useful information about the training data and the causal mechanism that generated it.
Our framework models (i) how a prediction from an ML model affects the beliefs of a rational Bayesian agent, and (ii) how this change in beliefs affects the estimation of the causal effect, the downstream decision for a new observation, and the subsequent outcome.  
In addition to the framework itself, we make three contributions. First, for the multivariate linear-Gaussian setting, we derive a tractable solution for the Bayesian inference problem introduced, i.e.~one in which the agent infers information about a population from an ML prediction.
Second, we experimentally identify conditions under which ML-DS is beneficial compared to unassisted decision making.
Third, we show that a single misaligned prior belief of the user can be sufficient for ML-DS to lead to worse downstream outcomes compared to no decision support \textit{even when the prognostic model is well-specified and the agent is perfectly rational}. 
Hence, even under ideal conditions, ML-DS can do more harm than good. \end{abstract}

\section{Introduction}

\paragraph{Background.} 

Prediction models are often used to support decision making in the following fashion: a model trained on past data produces a forecast for a future outcome of interest, which an agent takes into account when deciding how to act.  This is sometimes called `predictive optimization' \citep{Wang_Kapoor_Barocas_Narayanan_2024}. In contrast to full automation, in high-risk settings such an approach is sometimes perceived as a way to leverage the power of machine learning while retaining human oversight.\footnote{See for instance Article 14 of the EU AI Act \citep{EU_AI_Act_2024}.} Our work focuses on models that are \emph{prognostic} (i.e. the prediction is about a future state that can be influenced by the agent's actions) and \emph{treatment-naive} (i.e. the action variable is not among the model's predictors). We refer to such decision support as Machine Learning Decision Support (ML-DS) henceforth.

Understanding the impact of adopting ML-DS is essential for at least two reasons. First, ML-DS is on the rise and has been deployed in a variety of fields including healthcare \citep{escobar2020automated}, education \citep{smith2012predictive}, policing \citep{mohler2015randomized}, the judicial system \citep{imai2023experimental}, and child welfare \citep{chouldechova2018case}. Second, allowing prediction models to influence decision making carries important risks, e.g. the incorporation of biased advice \citep{obermeyer2019dissecting}. %
Despite this, how agents incorporate ML predictions into their model of the world and how this influences their actions has remained underexplored. We next present two scenarios where ML-DS interacts with users' priors in subtle ways.

\paragraph{Motivating example 1: Effect of the agent's prior beliefs.}
Consider the situation of a clinician using ML-DS when deciding how to treat a cancer patient drawn from the same population the ML-DS model was trained on. 
The ML-DS model predicts that the patient has 
only two months left to live.
\begin{wrapfigure}[21]{r}{0.58\textwidth}
    \centering
    \includegraphics[width=\linewidth]{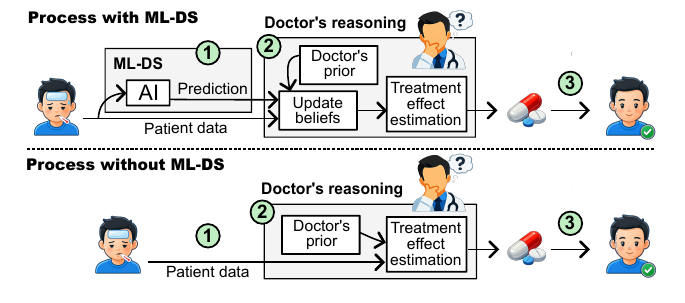}
    \caption{The flow of the framework for a clinical use case, with and without ML-DS. A new patient arrives for treatment, and the patient's data (1), potentially with the output of ML-DS, is observed by the agent, e.g. a doctor. Next,  the agent estimates the effect of treatment and makes a treatment decision (2). Without ML-DS, this estimation is influenced by the agent's priors and patient data. With ML-DS, the agent uses these plus the ML prediction to update their beliefs about the population in a Bayesian fashion. The treatment and the patient's attributes determine the outcome (3).}
    \label{fig:flowchart}
\end{wrapfigure}
We can imagine two lines of thinking, depending on the clinician's prior beliefs about the training data. If the model was trained on data from patients that received minimal treatment, a very heavy treatment may extend the survival time. On the other hand, if the model was trained on data from patients that received the heaviest treatment, it is best to put the patient on palliative care to avoid unnecessary suffering.

This example shows that in the presence of the same information from ML-DS, \textit{different prior assumptions about the model and the data it was trained on can lead to opposite actions}.   
This underscores the importance of an agent's priors in determining the effectiveness of ML-DS,\footnote{Setting the agents' prior beliefs by proper training %
is indeed an important part of RCT protocols that is not always adhered to \citep{perepletchikova2007treatment}. When it is not possible to create a common baseline of beliefs and behavior, a two-stage randomization might be required, both on the providers of the intervention and the subjects \citep{borm2005pseudo}.}
a point that was argued for in previous literature (see e.g., \cite{van2025risks}) but has not been studied quantitatively.

\paragraph{Motivating example 2: Agent learning from ML-DS.} \label{sec:motivating}
Consider a variation of the previous scenario involving a patient with a similar prognosis. Before seeing the model's prediction, the clinician believed (1) that similar patients ended up surviving 6 to 12 months when treated, and 
(2) that the training data of the model was from treated patients. Seeing a prediction of two months of survival, this set of beliefs is challenged. 
The clinician can then revise their beliefs in at least two different ways. First, they can conclude the training data in fact included some untreated patients, and that the current patient might be similar to one of the untreated patients. In this case, treatment might still be beneficial. Second, they might conclude they were wrong about the effect of the treatment. In this case, the treatment must be less beneficial than previously thought by the clinician. Therefore it might be best to abstain from treatment.
Hence \textit{the way in which ML-DS changes the agent's beliefs will influence the action of the agent}. Capturing this phenomenon means solving a challenging learning problem where the ML prediction is used as proxy information about the causal process the agent wants to intervene on.

\paragraph{Contributions.}
These examples showcase the importance of an agent's belief-updating process and prior beliefs for the impact of ML-DS. In this paper, we present a computational framework for ML-DS, \textit{the 2-Step Agent}, which includes a Bayesian step to incorporate the prediction by the ML model into the agent's model of the world, and a causal inference step in which the agent draws the conclusion on which action is best to take. This is schematically depicted in Fig.~\ref{fig:flowchart} (top half) for the clinical example. 
In contrast to standard prediction-then-decision approaches, the distinctive feature of our framework is that the agent treats the ML-DS prediction as a source of information about the training population and the latent causal mechanism, and uses this to update their beliefs about treatment effects before taking an action.
Such an update is a challenging inference through a set of latent exchangeable variables in the predictive model's training data. 

We make three further contributions. First, given the difficulty of solving the Bayesian inference problem we introduced, we prove that for the multivariate linear-Gaussian setting with a linear regression ML-DS model, it is possible to reduce the exchangeable variables to two auxiliary random variables, reducing computational cost and improving numerical stability. Second, we identify experimentally some settings in which learning from ML-DS leads to better effect estimates and better outcomes compared to not using ML-DS. Finally, we show that, even if the agent is fully rational and the model is well-specified, a single misaligned prior can be sufficient to derail learning from ML-DS and make the agent take suboptimal actions, resulting in worse outcomes. %

While our running example is medical, the results are general and apply to every field where treatment-naive prognostic prediction models are used for decision support. Our simulations are only concerned with learning from non-causal prediction models. ML-DS with causal models is left for future work. 
The code and experiments are available open-source at \url{https://github.com/thelogicalgrammar/CapturingUsersPredictionModels}.

\section{The 2-Step Agent framework}
\label{sec:framework}

In this section, we give a formal description of the 2-Step Agent. We introduce the process that generates the training data for the predictive model in Sec.~\ref{sec:setup}. We then formalize the first step consisting of a Bayesian agent updating their beliefs about the world given a prediction from the ML-DS model for a new patient, in Sec.~\ref{sec:firststep}. Sec.~\ref{sec:secondstep} defines the second step, where the Bayesian agent uses the updated beliefs about the population to decide on a treatment for the new patient. In Sec.~\ref{sec:quantifyingeffects} we combine these steps to determine the effect of introducing ML-DS, comparing outcomes with and without decision support. Sec.~\ref{sec:theorem} reports our tractability result.

\subsection{The setup: Historical data and predictive model}\label{sec:setup}

We first define a causal model of the domain. Let $A$ be a variable denoting an intervention, such as the administration of a treatment. The treatment variable may be continuous or categorical, but to simplify the notation, we will present the formulas with two options for $A$. Let $Y$ be a variable denoting a future outcome of interest affected by $A$. In our example, the treatment could be a fixed dose of chemotherapy and the outcome the months of survival, so in this case a higher outcome is better. We use $X \in \mathbb{R}^k$ to refer to the variables involved in the causal mechanisms governing the treatment and outcome.

\begin{definition}[Historical data generation]
    We assume $Hist = (\mathbf{S}, P_{\mathbf{N}})$ to be the minimal structural causal model (SCM) describing the causal mechanisms producing $A$ and $Y$, where $\mathbf{S}$ is a set of assignments defining each variable in terms of its parents and $P_{\mathbf{N}}$ is a joint distribution over the noise variables. We use $X$ to denote endogenous (non-noise) variables involved in the SCM that are different from $A$ and $Y$.
\end{definition}

An example of this SCM is depicted in Fig.~\ref{fig:agent_model}(a). The SCM $Hist$ stands for the process in the world that generated the training data. Note that the assignment determining $A$ might be due to a specific intervention protocol, e.g., the guidelines for administering the treatment that were in use in that particular time window. The SCM determines a joint distribution $P^{Hist}$ over variables $Y, A, X$.

\begin{definition}[Training data and prediction model]\label{def:data_predmodel}
    The dataset $\mathcal{D}_{Hist} = \{(x_1, a_1, y_1), \dots, (x_n, a_n, y_n)\}$ that is generated from $P^{Hist}$ is composed of tuples containing relevant characteristics, treatment, and outcome. The dataset contains $n$ instances. This dataset is used to train a prediction model $f: X \to Y$ to estimate $\mathbb{E}^{Hist}[Y\mid X]$.

\end{definition}

\noindent We call $f$ \emph{treatment-naive} since $A$ is not among its inputs, and \emph{well-specified} if its model class contains $\mathbb{E}^{Hist}[Y\mid X]$. Our framework does not currently extend to predictions of the treatment-dependent estimate $\mathbb{E}^{Hist}[Y\mid A, X]$ or the interventional estimate $\mathbb{E}^{Hist}[Y\mid do(A), X]$; these are left for future work.

\begin{wrapfigure}[24]{t}{0.6502\textwidth}
\centering
\vspace{0.04cm}
\begin{minipage}[c]{0.45\linewidth}
\centering
    \begin{subfigure}[t]{\linewidth}
    \centering
    \begin{tikzpicture}[scale=0.55, transform shape]
        \node[latent, xshift=-4cm] (Nx) {$N_{X}$};
        \node[latent, right=of Nx] (Ny) {$N_Y$};
        \node[latent, right=of Ny] (Na) {$N_A$};
        \node[latent, rectangle, below=of Nx] (X) {$X$};
        \node[latent, rectangle, below=of Na] (A) {$A$};
        \node[latent, rectangle, below=of Ny, yshift=-1.0cm] (Y) {$Y$};
        \edge {Nx} {X}; \edge {Na} {A}; \edge {Ny} {Y};
        \edge {X} {A}; \edge {A} {Y}; \edge {X} {Y};
    \end{tikzpicture}
    \caption{$Hist$}
    \end{subfigure}

    \vspace{0.8em}

    \setcounter{subfigure}{2}%
    \begin{subfigure}[t]{\linewidth}
    \centering
    \begin{tikzpicture}[scale=0.55, transform shape]
        \node[latent, xshift=-4cm] (Nx) {$N_{X}$};
        \node[latent, right=of Nx] (Ny) {$N_Y$};
        \node[latent, right=of Ny] (Na) {$N_A$};
        \node[latent, rectangle, below=of Nx] (X) {$X$};
        \node[latent, rectangle, below=of Na] (A) {$A$};
        \node[latent, rectangle, below=of Ny, yshift=-1.0cm] (Y) {$Y$};
        \node[latent, right=of Y] (E) {$N_E$};
        \edge {Nx} {X}; \edge {Na} {A}; \edge {Ny} {Y};
        \edge {X} {A}; \edge {A} {Y}; \edge {X} {Y};
        \edge {E} {Y};
    \end{tikzpicture}
    \caption{$\mathcal{A}_{Inf}$}
    \end{subfigure}
\end{minipage}%
\hfill
\begin{minipage}[c]{0.55\linewidth}
\centering
    \setcounter{subfigure}{1}%
    \begin{subfigure}[t]{\linewidth}
    \centering
    \begin{tikzpicture}[scale=0.55, transform shape]
    \tikzset{
        plate caption/.style={caption, node distance=0, inner sep=0pt, below right=5pt and 0pt of #1.south west}
        }
        \node[latent, xshift=-4cm] (priorsX) {$\alpha_{X}$};
        \node[latent, right=of priorsX] (priorsY) {$\alpha_Y$};
        \node[latent, right=of priorsY] (priorsA) {$\alpha_A$};
        \node[latent, below=of priorsX] (Nx) {$N_{Xi}$};
        \node[latent, below=of priorsY] (Ny) {$N_{Yi}$};
        \node[latent, below=of priorsA] (Na) {$N_{Ai}$};
        \node[latent, rectangle, below=of Nx] (X) {$X_i$};
        \node[latent, rectangle, below=of Na] (A) {$A_i$};
        \node[latent, rectangle, below=of Ny, yshift=-1.0cm] (Y) {$Y_i$};
        \node[latent, below right=1cm and 1cm of Y] (E) {$N_E$};
        \node[latent, rectangle, below=1.8cm of Y] (M) {$\theta$};
        \node[obs, right=of M, xshift=1.2cm] (Nm) {$N_{\theta}$};
        \node[obs, left=of Y, xshift=-2cm] (Xprime) {$X^{o}$};
        \node[obs, rectangle, below=of M] (Pred) {$Pred^{o}$};
        \plate {plate1} {(Nx)(Ny)(Na)(X)(A)(Y)} {$i=1,\ldots,n$};
        \edge {priorsX} {Nx}; \edge {priorsY} {Ny}; \edge {priorsA} {Na};
        \edge {Nx} {X}; \edge {priorsX} {Xprime};
        \edge {Na} {A}; \edge {Ny} {Y};
        \edge {X} {A}; \edge {A} {Y}; \edge {X} {Y};
        \edge {X,Y,A} {M}; \edge {Nm} {M};
        \edge {Xprime} {Pred}
        \edge {E} {Y}; \edge {M} {Pred};
    \end{tikzpicture}
    \caption{$\mathcal{A}_{Hist}$}
    \end{subfigure}
\end{minipage}
\caption{The graphical models of the framework with stochastic variables depicted as circles and deterministic variables as rectangles. Observed variables are shaded, unobserved ones are unshaded. Model (a) represents the true data-generating process,
model (b) represents the agent's beliefs of how the ML-DS model was trained, and %
model (c) allows the agent to infer the treatment effect of new data points given the beliefs. 
}
\label{fig:agent_model}
\end{wrapfigure}
\subsection{The first step: Bayesian update of the agent's beliefs}\label{sec:firststep}
We next introduce an agent $\mathcal{A}$ who decides on the intervention for new patients with the help of ML-DS. We assume this agent does not observe the training data or the model directly. 
This agent uses their own model of the world, capturing not only what they believe about the data-generating process $Hist$, but also what they believe concerning the prediction model $f$ and the training data $\mathcal{D}_{Hist}$.\footnote{Recent guidelines ask authors to ``give details of any treatments received, and how they were handled during model development or evaluation'' \citep{collins2024tripod+}, which would limit the agent's uncertainty about this aspect of the training data. Our framework can include such settings by restricting the uncertainty on this dimension.} 
The agent updates their world model using the ML-DS prediction, exploiting the latter as a piece of compressed information about the causal mechanism.
We model an agent who is aware of the model documentation containing the model \textit{class}, the model \textit{signature} (treatment-naive and included covariates), and the size of the training data. 

\begin{definition}[Beliefs of the agent]
    The Bayesian network defining the agent's joint distribution is denoted $\mathcal{A}_{Hist}$ (Fig.~\ref{fig:agent_model}(b)). The distribution is over the following variables:
    \begin{enumerate}[leftmargin=3em]
        \item $\alpha_{X}, \alpha_Y, \alpha_A$: These are the variables fixing the population-level distributions, e.g. over the patient characteristics ($\alpha_{X}$), the noise in the outcome ($\alpha_Y$), and the noise in the treatment ($\alpha_A$). Depending on how population-level uncertainty is parameterized, these can be vectors, for instance $\alpha_Y$ could be a pair of mean and variance if the noise on $Y$ is Gaussian. 
        \item $N_{Xi}, N_{Yi}, N_{Ai}$: Within the Bayesian network, the historical data is represented via a plate. The $\alpha$ variables are outside the plate, capturing the fact that they are population-level distributions. $N_{Xi}, N_{Yi}, N_{Ai}$ (with $i=1,\ldots,n$) are historical samples from these population-level distributions, which $\mathcal{A}$ cannot observe directly. These encode the stochastic components of each patient that the predictive model was trained on.
        \item $A_i, Y_i, X_i$: The stochastic components $N_{Xi}, N_{Yi}, N_{Ai}$ are transformed deterministically into the specific patient characteristic ($X_i$), treatment ($A_i$) and outcome ($Y_i$), which become a training data point for the predictive model.  All variables $Y_i$ share the noise term $N_E$.
        \item $N_E$: This variable denotes the treatment effect. Contrary to the other noise variables, this is not sampled for each plate repetition but only once, so it remains the same across the plate.
        \item $\theta$: The prediction model's parameters, obtained deterministically from the training data in the plate and possibly some additional known parameter $N_{\theta}$ which encodes e.g. the hyperparameter settings or initialization values. $\theta$ has as parents all the non-noise variables in the plate, representing the fact that it is trained on the whole data set.
        \item $X^o$: Observed variables for a new patient, for which the agent will make a treatment decision. This vector
        is sampled from the same population distribution as $X$ (for simplicity we skip the noise variable since it is identical to $X^o$). 
        \item $Pred^o$: The observed prediction for $X^o$. This is derived deterministically from $X^o$ and $\theta$.
    \end{enumerate}
\end{definition}

The Bayesian network is shown in Fig.~\ref{fig:agent_model}(b). To simplify presentation we depict the submodel of $\mathcal{A}_{Hist}$ in the plate as identical to $Hist$, meaning that the agent knows the DAG generating the data. This does not have to be the case and the framework is general enough to allow these two structures to diverge.

\begin{definition}[Agent's priors]
    The agent is assumed to have as prior a joint distribution over all variables in the model, which depends on the unconditional distribution $P^{\mathcal{A}_{Hist}}(\alpha_X, \alpha_Y, \alpha_A, N_E, N_\theta)$.
\end{definition}

The agent's prior beliefs about the data- and model-generating processes are thus encoded in $\mathcal{A}_{Hist}$ and the parameter sets $\alpha_X$, $\alpha_Y$, $\alpha_A$, $N_E$, and $N_\theta$. The set $\alpha_X$ contains all parameters necessary for the noise distribution of $N_X$ etc.
We assume the agent has no uncertainty on $N_\theta$
since they have access to the model documentation, and therefore we omit this prior.

\begin{definition}[Agent belief update]
    Given $X^o$ and $Pred^o$ the agent does a Bayesian update on the prior $P^{\mathcal{A}_{Hist}}(\alpha_{X}, \alpha_Y, \alpha_A, N_E)$, obtaining a posterior distribution $P^{\mathcal{A}_{Hist}}(\alpha_{X}, \alpha_Y, \alpha_A, N_E \mid X^o, Pred^o)$. 
    To simplify exposition, we marginalize over uncertainty in the plate variables ($N_{Xi}, N_{Yi}, N_{Ai},X_i, Y_i, A_i$ with $1\leq i\leq n$) and $\theta$.
\end{definition}

The uncertainty on $N_E$ encodes the agent's beliefs concerning the effect of the intervention.
By letting the mechanism be influenced by $N_E$, we can treat the posterior as encoding a distribution over SCMs, each with its own fixed assignment for $Y$ (and thus each with its own treatment effect). Other uncertainties over the SCM's parameters can be added as variables outside the plate like $N_E$; we leave these out to simplify exposition.

\subsection{The second step: deciding on the best action after belief update}\label{sec:secondstep}

In the second step, the agent estimates the effect of the treatment on the new patient using their updated beliefs from step 1. 

\begin{definition}[Agent model for treatment effect estimation, $\mathcal{A}_{Inf}$]
    This model contains variables  $N_E, N_{X}, N_A$, $N_Y, X, A, Y$, arranged as the similarly-named variables in the sub-model of $\mathcal{A}_{Hist}$. This model is used for inference on new instances and we denote it as $\mathcal{A}_{Inf}$.
\end{definition}

This second agent model is depicted in Fig.~\ref{fig:agent_model}(c). The crucial difference between this model and a regular SCM is that \textit{the distributional parameters of the noise variables are \emph{not} assumed to be independent but rather they are initialized with the joint posterior} obtained in the Bayesian update. The agent is thus drawing an inference over a distribution of possible SCMs.

\begin{definition}[Decision rule]
\label{def:decision_rule}
    To decide on treatment, the agent sets the joint distribution of the inference model drawing from the posterior learned in the previous step:
    $$
        \psi \sim
        P^{\mathcal{A}_{Hist}}(\alpha_{X}, \alpha_Y, \alpha_A, N_E \mid X^o, Pred^o) := \Pi
    $$
    Each realization of $\psi$ fixes the distribution of the noises and structural equations, specifying a different SCM $\mathcal{A}_{Inf}^{\psi}$. Following this, the treatment $a_1$ is given if the Conditional Average Treatment Effect (CATE) for the new patient provides enough benefit over treatment $a_2$ in expectation across SCMs, i.e. 
    $\mathbb{E}^{\Pi}[\text{CATE}^{\psi}]> \tau$
    where $\tau$ is a threshold set by the agent and CATE$^\psi$ is calculated via interventional expectations in $\mathcal{A}_{Inf}^{\psi}$ 
    $$\text{CATE}^{\psi} = \mathbb{E}^{\mathcal{A}_{Inf}^\psi}\big[Y \mid X = X^o, do(A=a_1)\big] - \mathbb{E}^{\mathcal{A}_{Inf}^\psi}\big[Y \mid X = X^o, do(A=a_2)\big].$$
\end{definition}

\subsection{Quantifying the effect of introducing decision support}\label{sec:quantifyingeffects}
Given this machinery, for a given ML-DS model and a set of agent priors, the treatment $A$ prescribed by the agent
for a new observation $X^o$ is deterministic both with and without ML-DS. In the former case (Fig.~\ref{fig:flowchart} top) the agent learns from ML-DS and chooses $A$ according to the updated beliefs. In the latter case, the agent does not change beliefs and the treatment effect is estimated based on the prior beliefs (Fig.~\ref{fig:flowchart} bottom). 
 We call the two treatments chosen with and without ML-DS $a^{ML}$ and $a^{\neg ML}$ respectively. The effect of ML-DS is then defined as
\begin{equation*}
   \mathbb{E}^{Hist}[Y\mid X=X^o, do(A = a^{ML})] - \mathbb{E}^{Hist}[Y\mid X=X^o, do(A = a^{\neg ML})]
\end{equation*}
namely as a difference of expected outcomes obtained by setting the actions to those chosen with and without ML-DS, both for the new patient $X=X^o$ and under the true generating process $Hist$.
In the experiments we report these two expected outcomes separately as distributions across simulation runs.

\subsection{Theoretical results on the tractability of the Bayesian inference}
\label{sec:theorem}
This framework is very general and allows for the simulation of many interesting learning dynamics. Yet, its implementation faces a tractability problem: approximate Bayesian inference struggles to propagate through a plate of completely exchangeable variables. 
We prove that this problem can in simple settings be solved by replacing the plate with two auxiliary random variables.

\begin{theorem}[Informal]
\label{thm:suff_stats}
Let $Hist$ be as depicted in Fig.~\ref{fig:agent_model}(a), where $X$ is a vector of $k$ independent covariates. Assume that the structural equations are linear, $n\geq k+1$, all variables are continuous, all noises are Gaussian, the agent knows the DAG and the ML-DS model is fitted by ordinary least squares with an intercept. Then, conditional on the population-level variables (the variables outside the plate), for a new patient independent of the training data, the likelihood of a single prediction given $X^o=x$ has an exact representation using two scalar auxiliary random variables.%
\end{theorem}

We refer the reader to Theorem~\ref{thm:single_prediction_collapsed} in Appendix~\ref{app:proof_thm_1} for the  formal statement and the full proof.
Corollary~\ref{corollary_no_intercept} gives the corresponding plate reduction for the no-intercept (misspecified) model, used as a robustness check. By collapsing the plate, the Bayesian update becomes tractable and we can compute the updated beliefs of the agent over the population-level parameters given the prediction obtained from the ML-DS model.
This unlocks the possibility to implement the framework (albeit in a simple setting): we can fully simulate what happens with and without decision support for a given agent $\mathcal{A}$ and an ML-DS model $f$ following the recipe outlined in the previous section.

\section{Experimental setup}
\label{sec:exp_setup}
The 2-Step Agent allows us to simulate the consequences of providing decision support.
Using this framework, we investigate \textit{how and how much decision support influences (i) agent's beliefs, (ii) treatment decisions, and (iii) downstream patient outcomes.} 
We run a series of simulations to see how beliefs, treatments, and outcomes are affected when a single prior is misaligned. We simulate all settings both with and without ML-DS to see the differences induced by ML-DS.

\paragraph{Case: univariate regression.}\label{sec:use_case}
We study what happens in the simple  setting when the ML-DS model is a well-specified linear regression model $f$ and the data-generating process is linear with one continuous covariate $X$, a continuous intervention $A$, and a continuous outcome $Y$. Higher values of $Y$ correspond to better outcomes. One can think of the motivating example and interpret $X$ as body weight, $A$ as dosage of chemotherapy, and $Y$ as months of survival. 
We use $n=1000$. Despite the continuous treatment, we assume the agent compares two treatment options, $A=10$ and $A=20$ (intuitively ten and twenty dosage units). 
Let $Hist$ have the DAG depicted in Fig.~\ref{fig:agent_model}(a) with a single $X$ variable. 

\paragraph{The model of the agent's beliefs, $\mathcal{A}_{Hist}$.} The model of the agent $\mathcal{A}_{Hist}$ encodes the agent's uncertainty about \textit{Hist}:
\begin{equation*}
\renewcommand{\arraystretch}{1.3}
\begin{array}{r@{\;}l@{\hspace{0.5em}}l@{\hspace{5.5em}}|r@{\;}l}
     N_{Xi} &\sim \mathcal{N}(\bm{\alpha_{X\mu}}, \bm{\alpha_{X\sigma}}^2) 
        & \rdelim\}{6}{2mm}[\scriptsize i=1,\ldots,n]
        & \theta_0, \theta_1 &= \mathop{\mathrm{argmin}}\limits_{\theta_0, \theta_1}(\sum\limits_{i=1}^n(\theta_0 + \theta_1 X_i- Y_i)^2) \\
     N_{Ai} &\sim \mathcal{N}(\bm{\alpha_{A \mu}}, \bm{\alpha_{A \sigma}}^2)
        & & X^o &\sim \mathcal{N}(\bm{\alpha_{X\mu}}, \bm{\alpha_{X\sigma}}^2) \\
     N_{Yi} &\sim \mathcal{N}(\bm{\alpha_{Y \mu}}, \bm{\alpha_{Y \sigma}}^2)
        & & Pred^o &= \theta_0 +\theta_1 X^o \\
     X_i & = N_{Xi} & & & \\
     A_i &= 0.125 \, X_i + N_{Ai} & & & \\
     Y_i &= 12 - 0.1 \, X_i + \bm{N_E} \, A_i + N_{Yi} & & &
\end{array}
\end{equation*}
where $\theta_0$ and $\theta_1$ are the intercept and slope of a univariate regression model, and the variables in bold font are the agent's priors for the population-level parameters.

The agent has uncertainty on several dimensions. For instance, $\alpha_{A \mu}$ encodes the belief of the agent on how much historical dosages of treatment deviated from the standard protocol, 
$\alpha_{X\mu}$ and $\alpha_{X\sigma}$ encode the belief of the agent on how the feature $X$ of the training data was distributed, and so forth. 
The distribution of $N_E$ captures the belief of the agent about the effectiveness of treatment $A$. Note that in this scenario the agent knows some of the SCM's parameters; this is for ease of exposition. Those parameters could also be replaced with random variables that the agent has uncertainty over (like $N_E$); the plate-reduction result covers this case under the assumptions stated in Appendix~\ref{app:proof_thm_1}.

\paragraph{Defining the agent.} We define a set of default priors with limited uncertainty to represent an agent that is "correct" in their beliefs.
\begin{align*}
    \alpha_{X\mu}&\sim \mathcal{N}(\textcolor{red}{80}, 0.1^2)&
     \alpha_{X\sigma}&\sim \mathcal{N}(10, 0.1^2) \mid \alpha_{X\sigma} \ge 0\\
     \alpha_{A\mu}&\sim \mathcal{N}(\textcolor{red}{2}, 1^2)&
     \alpha_{A\sigma}&\sim \mathcal{N}(1, 0.1^2) \mid \alpha_{A\sigma} \ge 0\\
     \alpha_{Y\mu}&\sim \mathcal{N}(\textcolor{red}{0}, 0.01^2)&
     \alpha_{Y\sigma}&\sim \mathcal{N}(0.1, 0.01^2) \mid \alpha_{Y\sigma} \ge 0\\
     N_E &\sim \mathcal{N}(\textcolor{red}{1}, 1^2)
\end{align*}
This agent essentially has the correct beliefs about the data-generating distribution, the historical treatment policy, outcome variance, and treatment effect. They only have some uncertainty represented by some variance on these parameters. In our experiments, we 
vary each of the parameters colored in red, i.e. the prior means of $\alpha_{X\mu}$, $\alpha_{A\mu}$, $\alpha_{Y\mu}$, and $N_E$
- one at a time - to simulate various ways in which the agent's beliefs may be incorrect. 
This only introduces misalignment in a mean of the agent's beliefs without affecting the beliefs on variability.

\paragraph{Reaching a decision.} 
Finally, the model of the agent used for inference, $\mathcal{A}_{Inf}$, is defined as in Fig.~\ref{fig:agent_model}(c).
The agent applies the decision rule in Definition~\ref{def:decision_rule} with $a_1=20$, $a_2=10$, and $\tau=5$. Hence, the agent administers a dosage of $20$ if the posterior expectation of CATE exceeds $5$ months, and a dosage of $10$ otherwise.

This completes the decision pipeline: given a patient $X^o$ with a prediction $Pred^o$ and the agent priors, we compute what the agent would do. 
For details of the implementation, see Appendix \ref{app:exp_setup}. The same experiments are extended to a multivariate setup with $k$ confounders, as well as to a univariate setup where the ML-DS model is a linear model without intercept (and is therefore \textit{not} well-specified); full details are in Appendices~\ref{app: multivariate_experiments} and \ref{app:no_intercept_experiments} respectively.

\section{Results}
\label{sec:results}
We begin by mentioning that our framework can capture the two motivating examples presented in the Introduction. What is clarified by this experiment is that the \textit{learning from ML-DS manifests itself through a correlation in the posterior of the agent}: the posterior correlation encodes the fact that the agent is entertaining multiple competing hypotheses to reconcile beliefs and observations. We expand on this in Appendix \ref{app:examples}.

\begin{figure}[t]
    \centering
    \includegraphics[width=\textwidth]{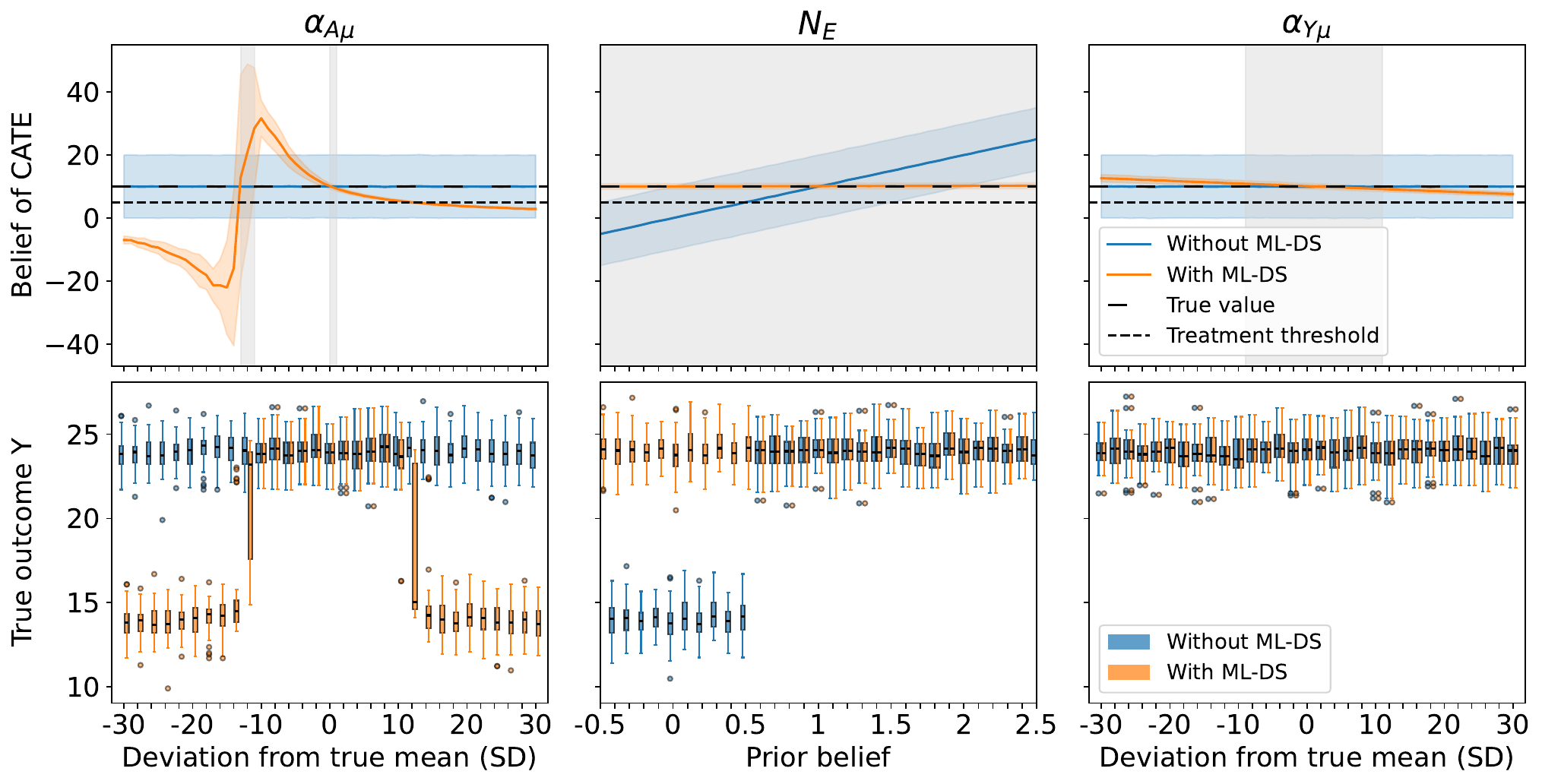}
    \caption{The effects of misaligned agent priors with ML-DS (in orange) and without (in blue) on the agent's beliefs (top row) and on the consequent downstream outcomes $Y$ (bottom row). 
    In the plots for $\alpha_{A\mu}$ and $\alpha_{Y\mu}$, the horizontal axes represent the deviation of the agent's prior belief from the true parameter in units of standard deviation, and the centerpoints in all plots correspond to aligned priors. 
    The orange and blue bands show the mean $\pm1$ SD of the agent's CATE distributions pooled across the 50 runs per prior setting, with and without ML-DS, respectively. Gray bands mark settings where the true CATE lies within the orange band. Boxplots summarize the 50 expected outcomes under the selected treatments and the true generating process $Hist$, showing the median and interquartile range (IQR), with whiskers extending to the most extreme values within 1.5 IQR of the box. The leftmost panel shows that misaligned beliefs can lead to ML-DS being harmful, while in the central panel ML-DS corrects misaligned beliefs and ends up being beneficial. 
    }
    \label{fig:misaligned_ANX_with_intercept}
    \vspace{-1cm}
\end{figure}

\paragraph{Univariate experiments with intercept.} We report the experimental results on two dimensions: 1) the agent's belief of treatment effect and 2) the expected downstream outcomes $Y$ under the true generating process $Hist$ after the agent has taken an action, visualized in the top and bottom rows of Fig.~\ref{fig:misaligned_ANX_with_intercept}. In each of these plots, what happens using ML-DS (orange series) is contrasted with what happens without (blue series) for different types of agents. The plots sweep a wide range of options for the priors in order to reveal interesting dynamics of the agent's beliefs.

The $N_E$ column in Fig.~\ref{fig:misaligned_ANX_with_intercept} shows what happens to the belief about CATE and the actual outcome $Y$ when the agent's belief about $N_E$ is misaligned. As expected, in the top row we can see that without decision support the prior mean of CATE (blue) is the prior mean of $N_E$ multiplied by the difference in treatment dosage. In the orange series, however, we see an interesting behavior: after learning from the interaction with ML-DS, the agent corrects their belief about CATE re-aligning it with the correct value. In other words, \textit{when all other beliefs are correct, the agent can learn from ML-DS to obtain better CATE estimates} and take better actions. The bottom row confirms that the orange series shows superior outcomes.

The column of Fig.~\ref{fig:misaligned_ANX_with_intercept} pertaining to $\alpha_{A\mu}$ shows what happens for the agents with different prior beliefs about the treatment protocol. In this scenario the prior CATE estimates are constant because the agent has a fixed belief in $N_E$. 
When interacting with ML-DS, the belief update changes CATE estimates quite a lot. This is because the agent is attempting to explain the gap between $Pred^o$ and their prior belief about it.
This is reflected in the outcomes in the boxplots, where the orange series is for some priors below the blue one. Thus, incorrect beliefs in $\alpha_{A\mu}$ can result in CATE estimates that lead to suboptimal actions and worse outcomes $Y$.
A similar pattern is observed in the column labeled $\alpha_{Y\mu}$,  concerning beliefs about the outcome distribution: while prior CATE and $Y$ estimates are relatively stable without ML-DS, CATE is affected when ML-DS is introduced, although in this specific example incorrect beliefs about $\alpha_{Y\mu}$ need to be very extreme to swing decisions. %
\begin{wrapfigure}[25]{r}{.68\linewidth}
    \centering    
    \vspace{0.8cm}
    \includegraphics[width=.68\textwidth]{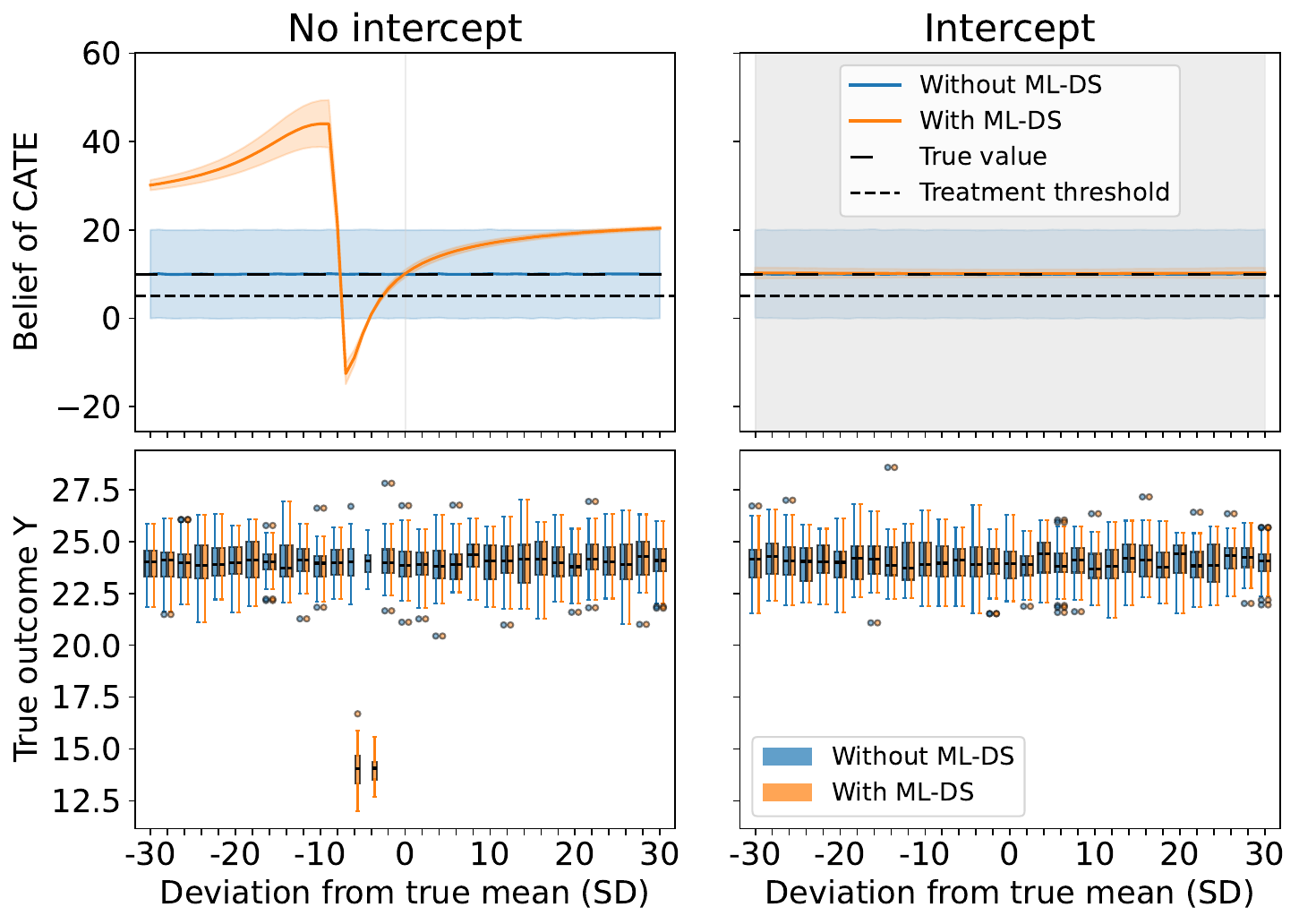}
    \caption{The effects of a misaligned agent prior $\alpha_{X\mu}$ about the distribution of covariate $X$, shown when the ML-DS model is misspecified (left) and well-specified (right), with bands and boxplots defined as in Fig.~\ref{fig:misaligned_ANX_with_intercept}. The misspecified model introduces a new option for incorrect beliefs to result in harmful ML-DS.
    }
    \label{fig:no_intercept_mu_X}
\end{wrapfigure}

\paragraph{Univariate experiment without intercept.} 
To explore what happens when the ML-DS model is misspecified, we ran an otherwise identical experiment, but with an ML-DS model without intercept. %
Fig.~\ref{fig:misaligned_ANY_no_intercept} shows that the same conclusions we discussed also hold  without intercept. 
The column pertaining to $N_E$ serves as an example for how our framework does not require the ML-DS model to be well-specified for useful learning to take place. Intuitively this is because the agent is merely using $Pred^o$ to learn about the generating process, rather than trying to incorporate this information in some more direct way.
Fig.~\ref{fig:no_intercept_mu_X}, however, displays a new failure mode: when the ML-DS model is not well-specified, misaligned beliefs about $\alpha_{X\mu}$ can also lead to harmful actions even with moderate misalignment.
See Appendices~\ref{app:no_intercept_experiments} and~\ref{app:no_intercept} for more details.

\paragraph{Multivariate regression with intercept.} 
Following the same pattern as for the well-specified univariate experiment, we ran an experiment where we varied the number of confounders $k$ in $X$ to see how it affects learning. 
We ran the experiment for $k \in \{2, 4, 8, 16, 32\}$, normalizing the confounders in a way that allowed us to isolate the effect of $k$ from e.g. increased variance. %
More details and the results are reported in Appendix~\ref{app: multivariate_experiments}.
This experiment reproduced the qualitative conclusions of the main experiment for $k>1$ and we found virtually no effect of $k$ on learning.

\section{Related work}

A large body of literature is concerned with the analysis of ML-DS from different angles.
A strand of literature focuses on how ML-DS is entangled with (changes in) the data-generating process. Works in this category study limitations of what can be learned from data generated from past policies \citep[e.g.][]{lakkaraju2017selective} or the dynamics that occur with alternating cycles of ML-DS deployment and re-training, as in the performative prediction literature \citep{perdomo2020performative, boeken2024evaluating}.  \cite{plecko2024causal} take a causal perspective on prediction-to-decision pipelines studying how biases in ML predictions amplify through downstream decisions. Their analysis treats the decision rule as a fixed mapping, where bias originates in the ML model. In contrast, our framework focuses on the decision maker, showing that a well-specified calibrated prediction paired with a misaligned prior can already be sufficient to worsen outcomes.

Another set of articles is concerned with the design and evaluation of ML-DS. Some lay out identifiability conditions for the effects of ML-DS on decisions \citep{imai2023experimental}, while others seek to optimize human-machine complementarity by design \citep{donahue2022human, mclaughlin2024designing}. Recent work in this direction studies prediction-set-based decision support \citep{detoni2024towards}, or evaluates the counterfactual harm that such systems may cause \citep{straitouri2024controlling}. Others evaluate the other side of the coin and ask what human input can add to artificial agents already reaching some level of optimality \citep{alur2024human, zhang2022can, stensrud2024optimal}. \cite{guo2024decision} propose a decision-theoretic framework for measuring AI reliance, explicitly adding a Bayesian component to the decision making, though one whose beliefs concern only the reliability of the prediction. Our framework addresses a different mechanism from all of the above: we explicitly model how the agent learns about the data-generating process from the ML prediction, and show that this learning can make decision support harmful even under near-ideal conditions.

A related line of work asks when different notions of calibration suffice for robust downstream decisions \citep{foster1997calibrated,zhao2021calibrating,kiyani2026robust}. These settings differ from ours in two crucial aspects. First, they are concerned with \textit{diagnostic} problems where the prediction target $Y$ is already set and not affected by actions (in contrast to our prognostic setup). Second, they assume a known payoff function for the available actions. We instead assume the user does \textit{not} have such a function and wants to estimate treatment effect with the help of ML-DS. It follows that calibration cannot solve the problem we presented. Calibration concerns agreement between predictions and observed outcomes under the historical policy. Even a perfectly calibrated predictor equal to $\mathbb{E}^{Hist}[Y\mid X]$ does not by itself determine how outcomes would change under an intervention and can still lead to harmful decisions under the wrong priors (see the example in Appendix~\ref{app:calibration_example}).

Other papers study how information flows between agents influencing their actions. 
In \cite{kamenica2011bayesian} a Sender of information influences a Receiver Bayesian agent in order to steer their action. What we add is that in our context the signal is an ML prediction produced from training data drawn from the same environment the agent wants to intervene on. This coupling yields the distinctive inference problem we study, i.e. updating through latent exchangeable training variables.

Another important strand of work focuses on inferring information about a population $P$ given predictions from a model trained on samples from $P$ in the context of confidentiality attacks.
This literature studies whether it is possible to use models' predictions (and other behavior) to recover the presence of a specific data point in so-called \textit{membership inference attacks} \citep{huMembershipInferenceAttacks2022, huangBayesianInferenceTraining2025}, or to infer statistical properties of training datasets (\textit{property inference attacks}) \citep{atenieseHackingSmartMachines2015,wangEavesdropCompositionProportion2019}. In contrast, our goal is to directly infer information about the underlying population and about the causal mechanism itself, i.e. the treatment effect.

\section{Discussion}

We have presented an explicit framework unpacking the interaction between an agent and ML-DS.
Our 2-Step Agent framework captures the effect of an agent's prior beliefs, modeling learning effects due to the interaction and tracking the agent's uncertainty. The framework allows for simulations comparing outcomes with and without ML-DS, quantifying the harm and benefit of ML-DS.

The framework defines a challenging Bayesian inference problem for which we derived  a tractable solution in the linear-Gaussian setting. This result allowed us to simulate the end-to-end effects of decision support for different types of agents, making this the first quantitative study of harmful effects of ML-DS with misaligned beliefs. 
In addition, we found that 
i) an incorrect belief about treatment effect is in some settings rectified by the interaction with ML-DS; 
ii) incorrect beliefs about the ML-DS model's training data may lead to harm, confirming arguments put forward in previous literature \citep{van2025risks}; and 
iii) incorrect beliefs about the distribution of covariates when the ML-DS model is not well-specified can lead to harm even in otherwise near-ideal conditions.

These findings highlight the sensitivity of ML-DS to prior beliefs when using a treatment-naive prognostic prediction model, and underscore the importance of providing full documentation and proper training to users of ML-DS. Where providers do not disclose such information, our results do not imply that ML-DS should be abandoned, but that its benefit cannot be taken for granted. Users' beliefs about the training data will determine whether it helps or harms. Finally, we remark that these results apply to \textit{any} field deploying such systems and to any Bayesian agent interacting with a treatment-naive prognostic ML-DS, be it human or artificial.

\paragraph{Towards real-world use.} The framework could be leveraged for pre-deployment audits, following these steps: (i) Elicit from prospective users, e.g.\ clinicians, priors over the quantities in $\mathcal{A}_{Hist}$: historical treatment policy, distributions of confounders, outcome and treatment effect. (ii) Simulate decisions and outcomes with and without ML-DS for each elicited prior. (iii) Identify users whose simulated outcomes worsen, and the specific belief responsible (as in Sec.~\ref{sec:results}). (iv) Adapt their training to fix said belief. This audit assumes approximately Bayesian updating and its validity for human users would need to be validated.

\paragraph{Limitations and future work.}

The experiments in this paper explored only part of the potential of the framework; they are limited to the multivariate linear-Gaussian case, with a treatment-naive ML-DS model, a single update and synthetic data. This was mostly due to the scope of Theorem \ref{thm:suff_stats}; further work will investigate whether similar tractability results can be obtained for more complex cases.

We assumed the  agent reasons in a Bayesian fashion. While this is not realistic with respect to human behavior, it is useful to make this assumption for three reasons. First, recent literature calls for formal computational cognitive models of the human decision maker as a necessary foundation for understanding human-AI complementarity, even when starting from a rational baseline \citep{steyvers2024three}.
Second, this assumption is a necessary first step to isolate the learning mechanism: if we had harmful outcomes caused by an agent with limited rationality, we would not know if it is due to the learning mechanism or just bounded rationality. Finally, this assumption is also interesting when considering how an artificial agent might learn from predictions of other ML models. 
From another point of view, our results indicate that harmful interactions with ML-DS due to misaligned priors cannot be fixed by making the agent more rational.

Our theory describes what happens after a single interaction of an agent with ML-DS and does not yet cover repeated interactions. 
In future work this line of research will be expanded across different dimensions, including i) different types of agents (e.g. agent with incorrect model of the world or different kinds of uncertainties), ii) more complex ML models for ML-DS, iii) cases with heterogeneity of treatment effect, iv) repeated rounds of learning for the same agent and v) using causal models for ML-DS.

\clearpage
\section*{Acknowledgements}
We would like to thank Wouter van Amsterdam, Nan van Geloven, Sagar N. Simha, Mohammad Azizmalayeri, and Juliette Ortholand for their feedback on draft versions of the paper. This publication was made possible through funding from the Sector Plan for Medicine and Health Sciences, supported by the Dutch Ministry of Education, Culture and Science (OCW). This work used the Dutch national e-infrastructure with the support of the SURF Cooperative using grant no. EINF-16946. We acknowledge ISCRA for awarding this project access to the LEONARDO supercomputer, owned by the EuroHPC Joint Undertaking, hosted by CINECA (Italy). Davide Tugnoli's PhD scholarship at the University of Trieste is funded by InSilicoTrials Technologies S.p.A.

\bibliographystyle{plainnat}
\bibliography{bibliography}

\newpage
\appendix
\onecolumn

\section{Motivating examples}\label{app:examples}
\begin{figure}[h!]
    \centering
  \includegraphics[width=0.3\linewidth]{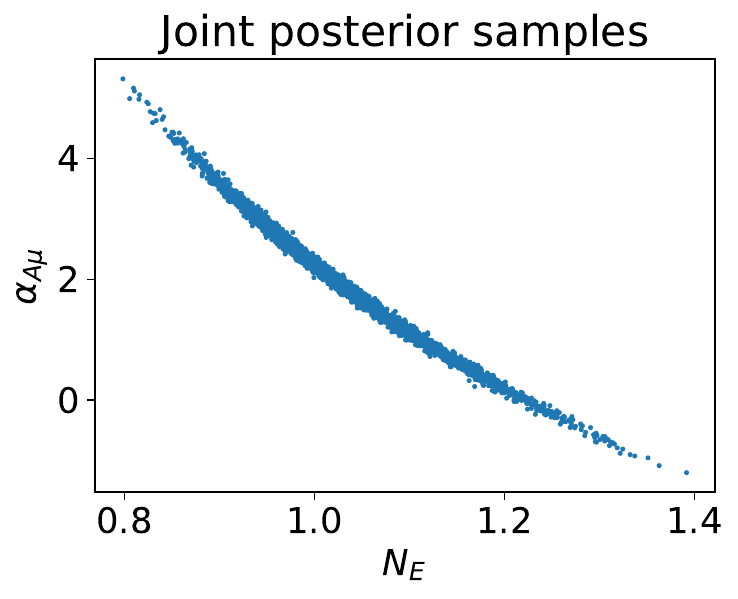}%
  \caption{Plot showing joint posterior samples of variables $N_E$ and $\alpha_{A\mu}$ in the posterior beliefs of the agent after update with ML-DS information. The negative correlation reflects that the agent entertains multiple possible explanations for the surprising prediction: either the treatment is less effective than expected, the historical dosages were higher than expected, or a scenario in between.}
  \label{fig:motivating_example_correlation}
\end{figure}

In this section we elaborate on how our framework and experimental setup can reconstruct and validate the intuition of the motivating examples.
In the first example, the belief of the agent on the past policy determines how the prediction is interpreted and what action is taken.
Indeed, in the simulation, when $X=80$, with a bad prognosis under the assumption of $\alpha_{A\mu} = 0$, the agent chooses a high dosage to improve the outcome ($A=20$). With a bad prognosis under the assumption of heavy treatment ($\alpha_{A\mu} = 20$), the agent chooses a low dosage to reduce over-treatment ($A=10$). In the second motivating example, new information from the prediction forces an agent to consider different options. 
In the model, this means that either belief in treatment effect ($N_E$) or in historical policy ($\alpha_{A\mu}$) is revised, which appears as a correlation in the posterior belief in these two variables in Fig.~\ref{fig:motivating_example_correlation}.

To recap: after observing a prediction that did not match expectation, the agent revises their beliefs, entertaining multiple hypotheses on how their beliefs should change to account for the situation. In this case, either the treatment effect is lower than anticipated and the dosage higher than assumed, or the other way around.

\section{Details on the experimental setup and implementation}
\label{app:exp_setup}

We report here all the details about the experiments, complementing Sec.~\ref{sec:exp_setup}.

\paragraph{Historical SCM.} 
Let the SCM $Hist$ be defined as:
\begin{align*}
     N_X &\sim \mathcal{N}(80, 10^2)\quad
     N_A \sim \mathcal{N}(2, 1^2)\quad
     N_Y \sim \mathcal{N}(0,0.1^2)\\
     X &:= N_X \\
     A &:= 0.125 \, X +N_A \\
     Y &:=  12-0.1 \, X + A + N_Y
\end{align*}

In this model, the treatment protocol is simply to give a dosage of treatment $A$ which is one eighth of the body weight $X$, and how well this dosing is followed is encoded by the noise term $N_A$. 
The mechanism of $Y$ clarifies that the covariate $X$ makes the outcome worse but the treatment improves the outcome. Of course, this equation is a great simplification of what happens in reality, but it is sufficient to demonstrate several important dynamics that emerge when introducing ML-DS.
Average treatment effect is positive in this scenario, as witnessed by the positive coefficient of $A$ in the equation for $Y$.

In general we denote the coefficient of $X$ in the treatment equation by $N_d$ and its coefficient in the outcome equation by $N_b$. In this univariate experiment they are fixed at $0.125$ and $-0.1$, respectively. In the multivariate experiments, with $k$ covariates, these coefficients are vectors (Appendix~\ref{app: multivariate_experiments}). More generally, they may be population-level random variables like $N_E$, as allowed by the theoretical results in Appendix~\ref{app:proof_thm_1}.

\paragraph{The model of the agent's beliefs.} 
The likelihood of the observed prediction $Pred^o$ is represented exactly via the auxiliary variables of Appendix~\ref{app:proof_thm_1}.

\paragraph{The model for the agent's effect estimation.} Finally, the model of the agent used for inference, $\mathcal{A}_{Inf}$, is defined as:
\begin{align*}
     N_{X} &\sim \mathcal{N}(\alpha_{X\mu}, \alpha_{X\sigma}^2) \quad
     N_{A} \sim \mathcal{N}(\alpha_{A \mu}, \alpha_{A \sigma}^2)\\
     N_{Y} &\sim \mathcal{N}(\alpha_{Y \mu}, \alpha_{Y \sigma}^2)\\
    X &:= N_{X}\\
     A &:= 0.125 \, X +N_{A} \\
     Y &:=  12 - 0.1 \, X + N_E \, A + N_{Y}
\end{align*}
\subsection{Implementation details}
\label{sec:implementation_details}
The experiments were run on two high performance computing clusters, Snellius (SURF, NL) and Leonardo (CINECA, IT). 
The main experiments (univariate ML-DS model with and without intercept) were run on Leonardo, for a total of 24,400 runs. Each run used 4 cores and 8 GB of RAM and took between 2 and 10 minutes.
The multivariate experiments were run on Snellius, for a total of 9,000 runs. Each run used 4 cores and 7 GB of RAM and took between 2 and 60 minutes.
We estimate that we used 20,000 CPU-hours in total for the final runs of the experiments.
The code for the simulations presented in this paper is based on PyMC \citep{pymc2023}.
We used the NUTS sampler to get 1000 samples for each of 4 chains, with 1000 tuning samples and a \texttt{target\_accept} parameter of 0.85. Diagnostics and visual inspection of the traces showed good mixing and no convergence issues (max $\hat{R}=1.01$, and a single divergence across all runs). 
For every value of a misaligned prior, we ran 50 such runs. Each run samples a new training set $\mathcal{D}_{Hist}$ from $Hist$, refits the ML-DS model on it, and samples a new patient $X^o$ from the true distribution.
Sanity checks on the Bayesian model are expanded upon in Appendix \ref{app:sanity_checks}. 
The plate reduction is derived in Appendix \ref{app:proof_thm_1}, with implementation notes in Appendix \ref{app:single_prediction_count}.

\section{Sanity checks on the Bayesian model}
\label{app:sanity_checks}
We have implemented a series of experiments to make sure that the framework is functioning as intended.
\begin{enumerate}
    \item We compare the posterior distributions with the explicit plate and with the reduced formulation given the same observations. We check if the posteriors of the two agents end up the same. This is repeated for different sizes of plate to control for potential effects of size.
    \item We run an update with many (1000) observations drawn from the real distribution of $X$ and check that the posterior of the agent converges to the real parameter $\alpha_{X\mu}$ in the generating process.
    \item We set one observation with $X$ equal to the mean of the agent's prior, with the priors on $A$ and $N_E$ very peaked, but with $Pred^o$ set much higher than the value $Y= N_b X + N_E A$. We also set $N_d$, the coefficient of $X$ in the structural equation of $A$, to zero. We observe in the posterior that there is now a correlation between the parameters of $N_E$ and $\alpha_{A \mu}$.
\end{enumerate}

\section{Results from univariate experiments with ML-DS model without intercept}
\label{app:no_intercept_experiments}

\begin{figure}
    \centering
    \includegraphics[width=\textwidth]{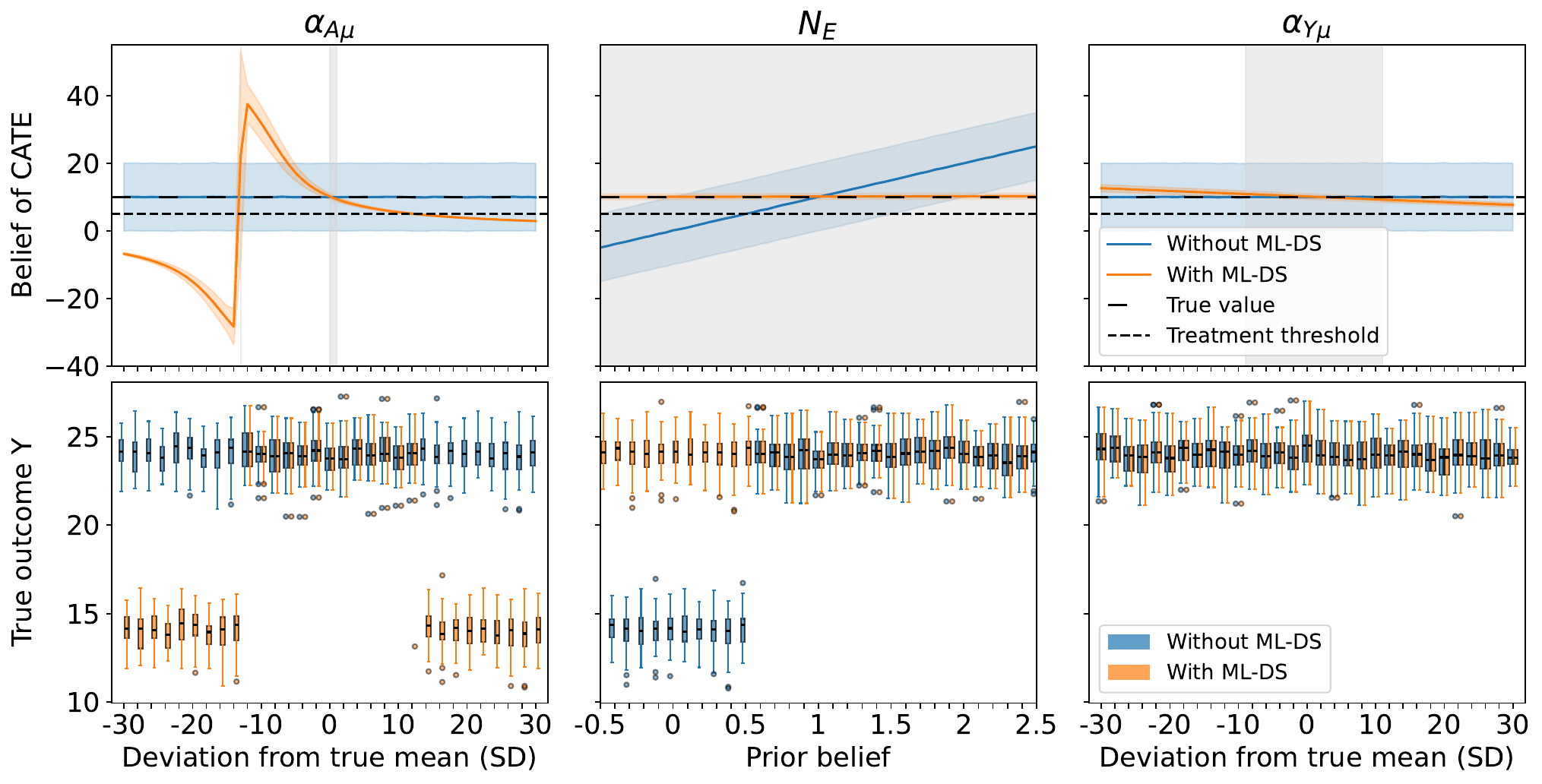}
    \caption{
    The effects of misaligned agent priors with ML-DS (in orange) and without (in blue) on the agent's beliefs (top row) and on the consequent downstream outcomes $Y$ (bottom row). 
    In contrast to Fig.~\ref{fig:misaligned_ANX_with_intercept}, the ML-DS model here is misspecified so that the regression model lacks intercept (thus it diverges from the true generating process).
    In the plots for $\alpha_{A\mu}$ and $\alpha_{Y\mu}$, the horizontal axes represent the deviation of the agent's prior belief from the true parameter in units of standard deviation, and the centerpoints in all plots correspond to aligned priors. 
    The orange and blue bands show the mean $\pm1$ SD of the agent's CATE distributions pooled across the 50 runs per prior setting, with and without ML-DS, respectively. Gray bands mark settings where the true CATE lies within the orange band. Boxplots summarize the 50 expected outcomes under the selected treatments and the true generating process $Hist$, showing the median and interquartile range (IQR), with whiskers extending to the most extreme values within 1.5 IQR of the box. Similarly to Fig.~\ref{fig:misaligned_ANX_with_intercept}, the leftmost plots show how a misaligned belief  $\alpha_{A\mu}$ can lead to the agent being more certain about flawed values of CATE with ML-DS, but this only leads to harmful treatment decisions when the belief drops below the treatment threshold. Similarly, the plots in the middle show how ML-DS can correct misaligned beliefs about the treatment effect over a wide range of priors which also results in improved treatment outcomes in some settings (where the blue line is below treatment threshold).
    The plot for misalignment of $\alpha_{X\mu}$ can be seen in Fig.~\ref{fig:no_intercept_mu_X}.}
    \label{fig:misaligned_ANY_no_intercept}
\end{figure}

We ran an experiment with an otherwise identical setup to the main experiment, but with an ML-DS model without intercept. In this setting, the ML-DS model is hence not well-specified. %
The results are presented in Fig.~\ref{fig:misaligned_ANY_no_intercept} and Fig.~\ref{fig:no_intercept_mu_X} (left).
The results for misaligned $\alpha_{A\mu}$, $N_E$, and $\alpha_{Y\mu}$ are almost identical to the results for the main experiment where the ML-DS model was well-specified. For $\alpha_{X\mu}$, however, we see very different dynamics in the two settings (Fig.~\ref{fig:no_intercept_mu_X}). 
While in the well-specified case the belief  $\alpha_{X\mu}$ was inconsequential, in this misspecified setting having a poorly aligned belief  $\alpha_{X\mu}$ can also lead to harm.

These experiments illustrate how our framework does not require the ML-DS model to be well-specified.
Column $N_E$ of Fig.~\ref{fig:misaligned_ANY_no_intercept} shows that useful learning can take place also when the ML-DS model is misspecified relative to both the true generating process $Hist$ and the generating process assumed by the agent.
The intuitive explanation for this is that our framework uses the prediction only to infer what the training data for a model that produced the prediction should have looked like, rather than e.g. attempting to incorporate the prediction in some more direct way.
This could in principle be done for any deterministic mathematical operation.

\section{Multivariate experiments}
\label{app: multivariate_experiments}

We investigated how the number of confounders $k$ in $X$ affects the belief updates, to see whether our findings from the univariate case extend also to the multivariate case. 
To isolate the effect of the number of confounders $k$, we kept certain properties of the model invariant. We specifically wanted the prior beliefs about CATE and its uncertainty to remain constant for all values of $k$ and all values of misaligned priors. 
Further, we wanted the agent's prior beliefs about the interventional outcomes and its uncertainties to remain constant for all values of $k$ and all values of misaligned priors.
The first one would guarantee that the agent's prior belief about the treatment effect is correct, whereas the second would guarantee that the agent's prior beliefs about outcomes are correct.
In a sense, these together would guarantee that the starting point of the agent is only different in terms of $k$.

We satisfied both invariances for misaligned beliefs  $\alpha_{X\mu}$ and $\alpha_{A\mu}$, and invariant prior CATE for misaligned beliefs  $\alpha_{Y\mu}$. For this experiment, $X$ is a vector with $k$ elements.
We used the same settings as in the main experiment (univariate ML-DS model with intercept) but with the following changes:
\begin{align}
    &(X_i)_1 \sim \mathcal{N}(\alpha_{X\mu}, \alpha^2_{X\sigma}) \\
    &(X_i)_j \sim \mathcal{N}(0, \alpha^2_{X\sigma}) \\
    &N_{b, 1} = -0.1 \\
    &N_{b, j} = \frac{-0.1}{\sqrt{k - 1}} \\
    &N_{d, 1} = 0.125 \\
    &N_{d, j} = \frac{0.125}{\sqrt{k - 1}}
\end{align}
for $j \in \{2, \ldots, k\}$.
This is equivalent to adjusting the variances of $(X_i)_j$ in a way that results in the variances of the inner products $N_b^\top X_i$ and $N_d^\top X_i$ being constant for all $k$. 
These experiments can only be run for $k \geq 2$, and we chose to run them for $k \in \{2, 4, 8, 16, 32\}$. We misaligned the beliefs about one variable at a time similarly to what was done in the main experiment, and of the confounders $X_i$ we only misaligned the beliefs about $(X_i)_1$ always keeping $(X_i)_j$ for $j > 1$ aligned.
Note that for $N_E$ the invariances cannot be satisfied as that variable directly encodes CATE. We chose to run the experiments with the same pattern for $N_E$ nevertheless.
As a side note, a slightly more natural setting would be to investigate how introducing more confounders without adjusting for the additional variance affects posterior beliefs. In such a setting, however, we could not have separated the effect of the number of confounders from the effect of increased variance.

\begin{figure}
    \centering
    \includegraphics[width=\textwidth]{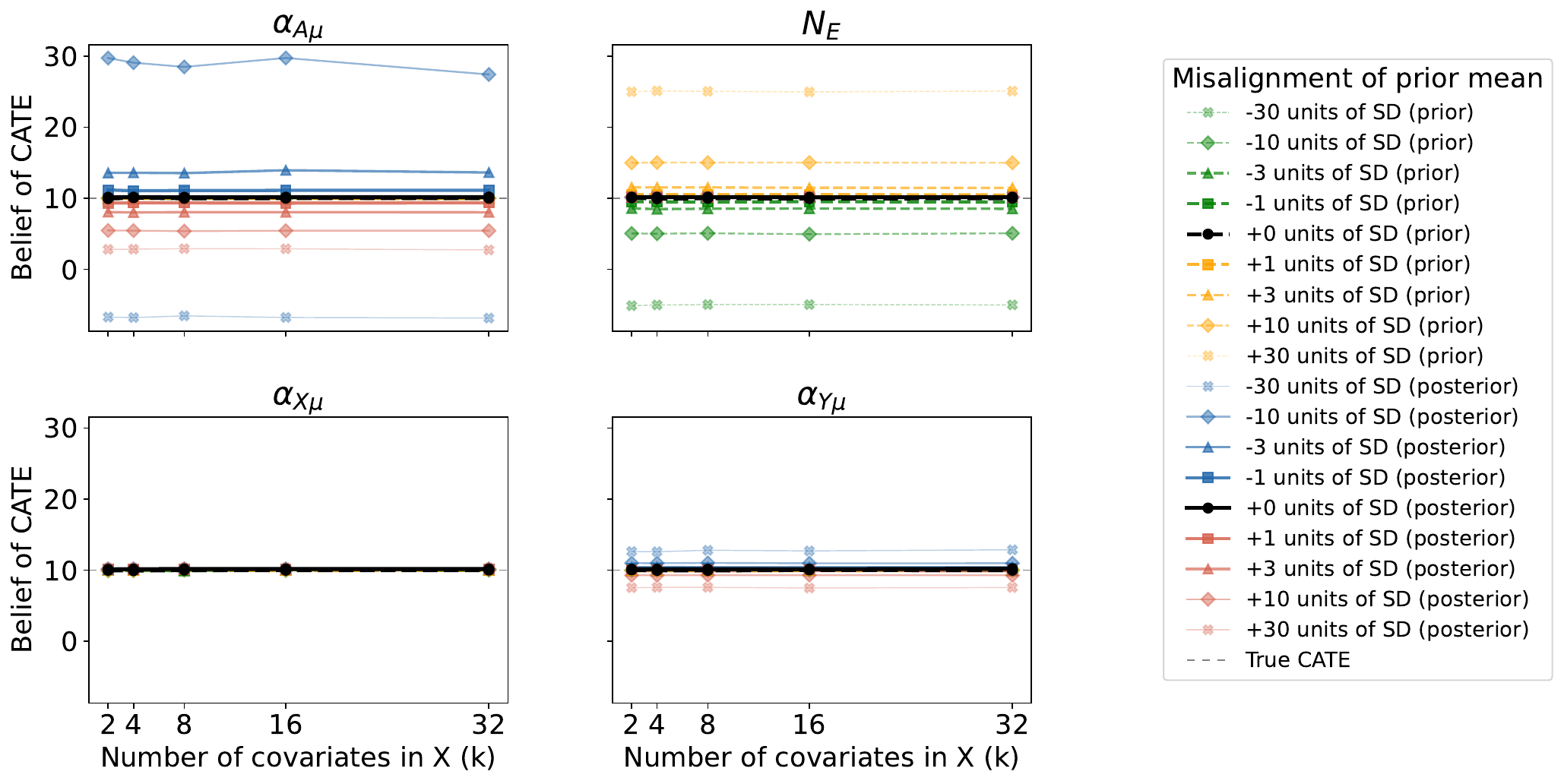}
    \caption{Number of covariates $k$ vs. prior and posterior beliefs about CATE for different misalignments of priors. For $N_E$, deviations of the prior mean are expressed in units of $0.05$.
    The experiment was constructed so that the prior expectation about CATE was correct despite the misaligned prior for $\alpha_{X\mu}$, $\alpha_{Y\mu}$, and $\alpha_{A\mu}$. For $N_E$ this could not be done due to the structure of $\mathcal{A}_{Inf}$.
    The findings from the main experiment are repeated in this setting and all of the results are essentially unaffected by $k$.
    Misalignment of $\alpha_{A\mu}$ and $\alpha_{Y\mu}$ can lead to the agent having less correct posterior beliefs about CATE, and misaligned beliefs about $N_E$ are largely corrected.
    Note that the lines that appear to be missing are simply masked by the thick black line.
    }
    \label{fig:dimensionality_means}
\end{figure}

\begin{figure}
    \centering
    \includegraphics[width=\textwidth]{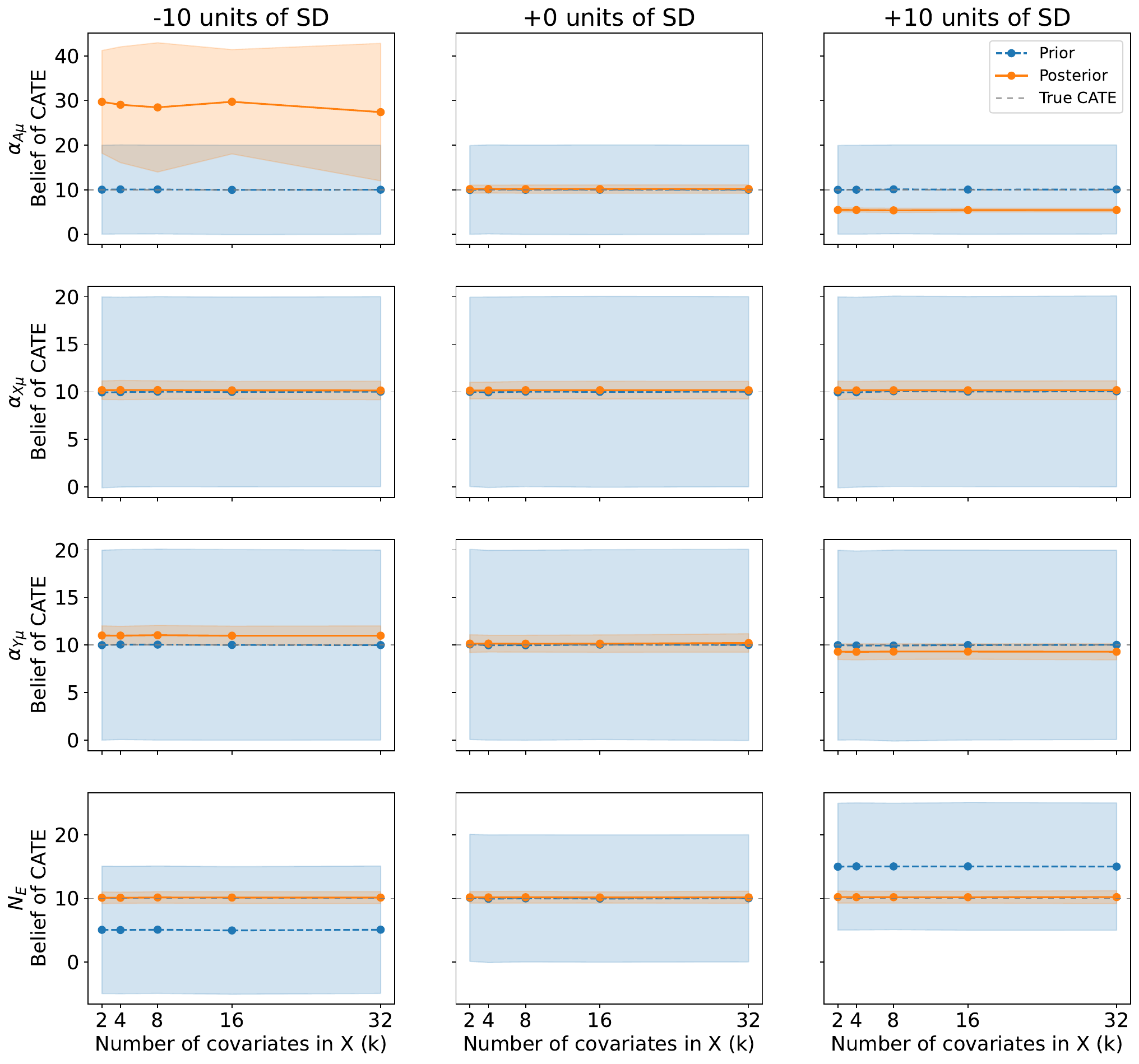}
    \caption{Prior (blue) and posterior (orange) beliefs about CATE vs. the number of covariates $k$. Lines show the mean and bands $\pm1$ SD of the agent's CATE distribution, pooled across the 50 runs per setting. For $N_E$, deviations of the prior mean are expressed in units of $0.05$.
    Both the posterior expectation of CATE and its uncertainty seem largely unrelated to $k$.
    The same effect with $N_E$ being corrected over a range of combinations seen in the main experiment is repeated here.}
    \label{fig:dimensionality_with_uncertainty}
\end{figure}

The results of these experiments are presented in Fig.~\ref{fig:dimensionality_means} and Fig.~\ref{fig:dimensionality_with_uncertainty}. 

Fig.~\ref{fig:dimensionality_means} shows the posterior mean of CATE. Due to how we set the experiment up (prior belief about CATE invariant), the prior expectation of CATE matched the true value in all settings except those with a misaligned prior mean for $N_E$ (dotted lines). Thus, to investigate how the beliefs were affected, it is sufficient to analyze the posterior values of CATE.
Given that the plots are essentially horizontal lines, we observe no discernible effect of the number of confounders $k$ on the belief update.

We further illustrate how the uncertainties about CATE were affected by $k$ in Fig.~\ref{fig:dimensionality_with_uncertainty}. 
The plots indicate no relation between the number of confounders $k$ and the posterior uncertainty of CATE.
We suspect that this is due to the plate size (number of observations the ML-DS model was trained on, $1000$) being much larger than $k$, which results in very little model uncertainty, but we have not tested this.
Also for $N_E$, we see the main results repeated. The plot shows how ML-DS largely corrects a misaligned belief about $N_E$ with $k$ having virtually no effect on the posterior.

\section{Plate reduction}
\label{app:proof_thm_1}

This appendix proves the plate-reduction result used for the Bayesian update in $\mathcal{A}_{Hist}$. We construct an exact representation of the conditional joint distribution of the fitted parameters $\theta$, replacing the $n$ latent training units in the plate with a small set of auxiliary variables. This representation preserves the likelihood of observations that depend on the training data only through $\theta$. Examples include the fitted coefficients themselves and one or more predictions for new patients who are independent of the training data given the population-level variables.

\begin{lemma}[Reduction of the plate to auxiliary random variables]
\label{thm:suff_stats_app}
Let the historical DAG be as depicted in Fig.~\ref{fig:agent_model}(a), where $X$ is a vector of $k$ independent covariates, and let $n$ be the number of data points in the historical training data. Assume that:
\begin{enumerate}
    \item the structural equations of the historical SCM are linear and without interaction terms;
    \item all variables are continuous;
    \item conditional on the population-level variables, patient-level noises are mutually independent across patients and variables and are Gaussian, with strictly positive covariate and outcome-noise variances;
    \item the ML model is fitted by ordinary least squares with an intercept, predicting $Y$ from $X$;
    \item $n\geq k+1$;
    \item the agent knows the historical DAG, the structural functional forms, the training sample size, the model signature and the fitting procedure.
\end{enumerate}
Then, conditional on the population-level variables, the joint distribution of the fitted parameters $\theta$ in $\mathcal{A}_{Hist}$ has an exact representation using $\frac{k^2+5k+2}{2}$ scalar auxiliary variables.
\end{lemma}

\begin{proof}
Throughout the proof we condition on the population-level variables
\[
N_b,\quad N_d,\quad N_E,\quad
\vec{\alpha}_{X\mu},\quad D_X,\quad
\alpha_{A\mu},\quad \alpha_{A\sigma},\quad
\alpha_{Y\mu},\quad \alpha_{Y\sigma},
\]
unless additional conditioning is stated explicitly. Uncertain population-level quantities are sampled once per realization of $\mathcal{A}_{Hist}$; known quantities are held fixed. In both cases they are shared by all patients in the plate.

The plate variables are the $n$ patient-level latent variables and observations,
\[
X_i,\quad A_i,\quad Y_i,\quad N_{Ai},\quad N_{Yi},
\qquad i=1,\ldots,n.
\]
The variables introduced below,
\[
s,\quad W,\quad g,\quad h,
\]
are auxiliary variables for the proof and for the reduced implementation. They are not new SCM variables. They replace the distributional role of the plate when computing the OLS coefficients.

The proof first rewrites the design matrix in standardized coordinates and splits the covariate plate into its mean, captured by $s$, and its centered variation, captured by $W$. It then folds the treatment equation into a single regression of $Y$ on $X$ and writes the fitted slope and intercept as functions of $(s,W)$ and two Gaussian noises $g$ and $h$. Finally, it shows that $s$, $W$, $g$ and $h$ are mutually independent.

\subsection{Multivariate SCM setup}
\label{app:thm_SCM}

For each historical patient $i=1,\ldots,n$, with covariate vector $X_i\in\mathbb{R}^k$, the agent's historical SCM inside the plate is
\begin{align}
X_i &= N_{Xi},
\label{eq:multi_scm_x}\\
A_i &= N_d^\top X_i + N_{Ai},
\label{eq:multi_scm_a}\\
Y_i &= N_b^\top X_i + N_E A_i + N_{Yi}.
\label{eq:multi_scm_y}
\end{align}
Here $N_b,N_d\in\mathbb{R}^k$ and $N_E\in\mathbb{R}$ are population-level variables, shared across the plate. The conditional derivation permits uncertainty about these coefficients, whereas the experiments fix $N_b=b$ and $N_d=d$. The univariate experiment uses $b=-0.1$ and $d=0.125$. In the multivariate experiment, the first covariate retains these coefficients, while the remaining covariates have the same coefficients scaled by $1/\sqrt{k-1}$. These fixed, coordinate-dependent coefficients are covered by Appendix~\ref{app:single_prediction_general}.

The noise distributions are
\begin{align}
(N_{Xi})_j &\sim \mathcal{N}(\alpha_{X\mu,j},\alpha_{X\sigma,j}^2),
\qquad j=1,\ldots,k,
\label{eq:noise_x}\\
N_{Ai} &\sim \mathcal{N}(\alpha_{A\mu},\alpha_{A\sigma}^2),
\label{eq:noise_a}\\
N_{Yi} &\sim \mathcal{N}(\alpha_{Y\mu},\alpha_{Y\sigma}^2).
\label{eq:noise_y}
\end{align}
All patient-level noises are mutually independent across patients and across variables. We assume $\alpha_{X\sigma,j}>0$, so the diagonal scaling matrix introduced below is invertible.

The explicit structural constant $12$ in the experimental outcome equation is absorbed into the noise mean in \eqref{eq:noise_y}: $\alpha_{Y\mu}^{\mathrm{appendix}}=12+\alpha_{Y\mu}^{\mathrm{main}}$. The prior mean is shifted by $12$, with its variance unchanged. This structural constant is distinct from the fitted OLS intercept.

\subsection{Standardization of the covariates}

We express the patient-level exogenous noises as their population means plus scaled independent standard Gaussian variables. This exposes the standard-normal matrix $Z$ from which the auxiliary variables $s$ and $W$ are built, and separates the population-level parameters $\vec\alpha_{X\mu}$ and $D_X$ from the standardized noise matrix $Z$.

Let
\[
\vec{\alpha}_{X\mu}
=
(\alpha_{X\mu,1},\ldots,\alpha_{X\mu,k})^\top
\in\mathbb{R}^k,
\qquad
D_X
=
\operatorname{diag}(\alpha_{X\sigma,1},\ldots,\alpha_{X\sigma,k})
\in\mathbb{R}^{k\times k}.
\]
Then each covariate vector can be written as
\[
X_i=\vec\alpha_{X\mu}+D_Xz_i,\qquad z_i\sim\mathcal N_k(0,I_k).
\]
The vectors $z_1,\ldots,z_n$ are independent. The diagonal matrix $D_X$ rescales the standardized covariates back to the population scale.

Similarly, write
\[
N_{Ai}
=
\alpha_{A\mu}+\alpha_{A\sigma}\xi_{A_i},
\qquad
N_{Yi}
=
\alpha_{Y\mu}+\alpha_{Y\sigma}\xi_{Y_i},
\]
where $\xi_{A_i}$ and $\xi_{Y_i}$ are i.i.d.\ $\mathcal{N}(0,1)$, independently of all $z_j$.

Let $Z\in\mathbb{R}^{n\times k}$ be the matrix whose $i$-th row is $z_i^\top$, and let $X\in\mathbb{R}^{n\times k}$ be the design matrix whose $i$-th row is $X_i^\top$. Then
\begin{equation}
X
=
\mathbf{1}\vec{\alpha}_{X\mu}^{\top}+ZD_X,
\label{eq:design_decomposition}
\end{equation}
where $\mathbf{1}\in\mathbb{R}^n$ is the vector of ones.

\subsection{Centering and projection matrices}

Since the model is fitted with an intercept, the slope depends on the covariates and outcome only through their deviations from the sample means. We introduce the two projections that carry out this split.

Define
\[
P=\frac{1}{n}\mathbf 1\mathbf 1^\top,\qquad M=I_n-P.
\]
The matrix $P$ projects onto the span of $\mathbf{1}$ and extracts sample means. The matrix $M$ projects onto the orthogonal complement of $\mathbf{1}$ and centers vectors. Both matrices are symmetric ($P^\top=P$ and $M^\top=M$) and idempotent ($P^2=P$ and $M^2=M$). They satisfy $P+M=I_n$, $PM=MP=0$ and $M\mathbf{1}=0$.

Let $Y=(Y_1,\ldots,Y_n)^\top\in\mathbb R^n$ be the outcome vector. For OLS with an intercept, the fitted slope vector $\theta_1\in\mathbb R^k$ and intercept $\theta_0\in\mathbb R$ are
\begin{equation}
\theta_1
=
(X^\top MX)^{-1}X^\top MY,
\qquad
\theta_0
=
\bar Y-\bar X^\top\theta_1,
\label{eq:ols_intercept}
\end{equation}
where
\[
\bar X=\frac{1}{n}X^\top\mathbf{1},
\qquad
\bar Y=\frac{1}{n}\mathbf{1}^\top Y.
\]

Using \eqref{eq:design_decomposition},
\[
MX
=
M(\mathbf{1}\vec{\alpha}_{X\mu}^{\top}+ZD_X)
=
MZD_X,
\]
because $M\mathbf{1}=0$. Hence
\begin{align}
X^\top MX
&=
(MX)^\top(MX)
\nonumber\\
&=
(MZD_X)^\top(MZD_X)
\nonumber\\
&=
D_XZ^\top MZD_X.
\label{eq:centered_gram_factorization}
\end{align}
The mean vector $\vec{\alpha}_{X\mu}$ disappears from the centered Gram matrix, so the slope denominator depends on the covariate plate only through its centered variation.

\subsection{The auxiliary variable \texorpdfstring{$s$}{s}}

The population covariate mean affects the fitted model through its intercept. By \eqref{eq:design_decomposition},
\begin{align*}
\bar X&=\frac1n X^\top\mathbf1\\
&=\vec{\alpha}_{X\mu}+\frac1n D_XZ^\top\mathbf1.
\end{align*}
We therefore define the $k$-vector
\begin{align*}
s&=Z^\top\mathbf1,\\
s&\sim\mathcal N_k(0,nI_k).
\end{align*}
Its distribution follows because $s$ is the sum of $n$ independent $\mathcal N_k(0,I_k)$ vectors. Thus
\begin{equation}
\bar X=\vec{\alpha}_{X\mu}+\frac1nD_Xs.
\label{eq:xbar_reduced}
\end{equation}

\subsection{The auxiliary Wishart matrix \texorpdfstring{$W$}{W}}

By \eqref{eq:centered_gram_factorization}, the slope denominator depends on the covariate plate through the centered cross-products of the standardized covariates. Define the $k\times k$ matrix $W=Z^\top MZ$, so that
\begin{equation}
X^\top MX=D_XWD_X.
\label{eq:cxx_reduced}
\end{equation}

To identify the distribution of $W$, let $U\in\mathbb{R}^{n\times(n-1)}$ have orthonormal columns spanning the subspace orthogonal to $\mathbf{1}$. Then
\begin{align*}
M&=UU^\top,\\
U^\top U&=I_{n-1}.
\end{align*}
Therefore
\begin{align*}
W&=Z^\top MZ\\
&=Z^\top UU^\top Z\\
&=(U^\top Z)^\top(U^\top Z).
\end{align*}
The columns of $Z$ are independent $\mathcal N_n(0,I_n)$ vectors, so the columns of $U^\top Z$ are independent $\mathcal N_{n-1}(0,I_{n-1})$ vectors, since $U^\top U=I_{n-1}$. Hence all entries of $U^\top Z$ are independent standard normals, and its rows are independent $\mathcal N_k(0,I_k)$ vectors. By the definition of the Wishart distribution \citep[Definition~3.1.3]{muirhead1982},
\[
W\sim\operatorname{Wishart}_k(I_k,n-1).
\]
The assumption $n\ge k+1$ gives $n-1\ge k$, so $W$ is positive definite almost surely \citep[Theorem~3.1.4]{muirhead1982}.

\subsection{Independence of \texorpdfstring{$s$}{s} and \texorpdfstring{$W$}{W}}

To specify the joint distribution of $(s,W)$, we now establish their independence.

The variable $s$ depends only on the projection of $Z$ onto the span of $\mathbf{1}$:
\begin{align*}
PZ&=\frac{1}{n}\mathbf{1}\mathbf{1}^\top Z\\
&=\frac{1}{n}\mathbf{1}s^\top.
\end{align*}
The matrix $W$ depends only on the centered projection:
\begin{align*}
W&=Z^\top MZ\\
&=(MZ)^\top(MZ).
\end{align*}
For each column $z$ of $Z$, the vectors $Pz$ and $Mz$ are jointly Gaussian with $\operatorname{Cov}(Pz,Mz)=PM^\top=PM=0$, hence independent. Since the columns of $Z$ are independent, $PZ$ and $MZ$ are independent. Therefore $s$, as a function of $PZ$, is independent of $W$, as a function of $MZ$:
\[
s\perp W.
\]
Together with the preceding subsection, this recovers the classical independence of the sample mean and the centered cross-product matrix of an i.i.d.\ Gaussian sample, the latter having a Wishart distribution with $n-1$ degrees of freedom \citep[Theorem~3.1.2]{muirhead1982}.

\subsection{Condensing \texorpdfstring{$A$}{A} into \texorpdfstring{$Y$}{Y}}

The fitted model regresses $Y$ on $X$, but in the SCM, part of $Y$'s dependence on $X$ is routed through $A$. We first collapse the two structural equations into one effective regression of $Y$ on $X$, whose slope, intercept, and error variance absorb the treatment channel.

Substitute \eqref{eq:multi_scm_a} into \eqref{eq:multi_scm_y}:
\begin{align*}
Y_i
&=
N_b^\top X_i
+
N_E(N_d^\top X_i+N_{Ai})
+
N_{Yi}
\\
&=
N_b^\top X_i
+
N_E N_d^\top X_i
+
N_E(\alpha_{A\mu}+\alpha_{A\sigma}\xi_{A_i})
+
\alpha_{Y\mu}+\alpha_{Y\sigma}\xi_{Y_i}
\\
&=
\underbrace{\alpha_{Y\mu}+N_E\alpha_{A\mu}}_{\beta_0}
+
\underbrace{(N_b+N_EN_d)^\top}_{\beta_1^\top}X_i
+
\underbrace{N_E\alpha_{A\sigma}\xi_{A_i}
+
\alpha_{Y\sigma}\xi_{Y_i}}_{\varepsilon_i}.
\end{align*}
Here $\varepsilon_i$ is the regression error. Because $\xi_{A_i}$ and $\xi_{Y_i}$ are independent standard normal variables, $\varepsilon_i\sim\mathcal{N}(0,\sigma^2)$, where we define
\[
\sigma^2=N_E^2\alpha_{A\sigma}^2+\alpha_{Y\sigma}^2.
\]
Since $\alpha_{Y\sigma}>0$ by assumption, also $\sigma>0$.
The regression errors are independent across $i$, and they are independent of all covariate noises $z_j$. In vector form,
\begin{equation}
Y
=
\beta_0\mathbf{1}+X\beta_1+\varepsilon,
\qquad
\varepsilon\sim\mathcal{N}_n(0,\sigma^2I_n),
\qquad
\varepsilon\perp Z.
\label{eq:y_condensed}
\end{equation}
Thus the treatment equation can be absorbed into an effective linear regression of $Y$ on $X$ with slope $\beta_1$, intercept $\beta_0$, and error variance $\sigma^2$.

\subsection{OLS slope with intercept}

Substitute \eqref{eq:y_condensed} into the slope formula in \eqref{eq:ols_intercept}:
\begin{align}
\theta_1
&=
(X^\top MX)^{-1}X^\top M(\beta_0\mathbf{1}+X\beta_1+\varepsilon)
\nonumber\\
&=
(X^\top MX)^{-1}
\left[
\beta_0X^\top M\mathbf{1}
+
X^\top MX\,\beta_1
+
X^\top M\varepsilon
\right]
\nonumber\\
&=
\beta_1+(X^\top MX)^{-1}X^\top M\varepsilon.
\label{eq:slope_noise_start}
\end{align}
The term $\beta_0X^\top M\mathbf{1}$ is zero because $M\mathbf{1}=0$. This is where centering removes the intercept from the slope calculation.

Conditionally on $Z$, the vector $X^\top M\varepsilon$ is a linear transformation of the Gaussian error vector $\varepsilon$. Since $\varepsilon\mid Z\sim\mathcal{N}_n(0,\sigma^2I_n)$,
\begin{align}
\mathbb{E}[X^\top M\varepsilon\mid Z]
&=
0,
\\
\operatorname{Cov}(X^\top M\varepsilon\mid Z)
&=
X^\top M(\sigma^2I_n)M^\top X
\nonumber\\
&=
\sigma^2X^\top MX
\nonumber\\
&=
\sigma^2D_XWD_X.
\end{align}
The conditional distribution depends on $Z$ only through $W$. Hence
\begin{equation}
X^\top M\varepsilon\mid W
\sim
\mathcal{N}_k(0,\sigma^2D_XWD_X).
\label{eq:xmeps_conditional}
\end{equation}

Let $W^{1/2}$ denote the symmetric positive definite square root of $W$. A valid covariance factor for the covariance in \eqref{eq:xmeps_conditional} is $\sigma D_XW^{1/2}$, because
\[
(\sigma D_XW^{1/2})(\sigma D_XW^{1/2})^\top
=
\sigma^2D_XWD_X.
\]
Define
\[
g=\left(\sigma D_XW^{1/2}\right)^{-1}X^\top M\varepsilon,
\]
so that
\begin{equation}
X^\top M\varepsilon=\sigma D_XW^{1/2}g.
\label{eq:g_definition}
\end{equation}
Conditionally on $Z$, the vector $g$ is a linear transformation of the Gaussian vector $X^\top M\varepsilon$, with mean $0$ and covariance
\[
\left(\sigma D_XW^{1/2}\right)^{-1}
\sigma^2D_XWD_X
\left(\sigma D_XW^{1/2}\right)^{-\top}
=
I_k.
\]
Hence $g\mid Z\sim\mathcal{N}_k(0,I_k)$ for almost every realization of $Z$. Because this conditional distribution does not depend on $Z$, the vector $g$ is independent of $Z$, and therefore independent of $(s,W)$, which are functions of $Z$:
\[
g\sim\mathcal{N}_k(0,I_k),
\qquad
g\perp(s,W).
\]

Substituting \eqref{eq:g_definition} and \eqref{eq:cxx_reduced} into \eqref{eq:slope_noise_start},
\begin{align*}
\theta_1
&=
\beta_1
+
(D_XWD_X)^{-1}\sigma D_XW^{1/2}g
\\
&=
\beta_1
+
\sigma D_X^{-1}W^{-1}D_X^{-1}D_XW^{1/2}g
\\
&=
\beta_1+\sigma D_X^{-1}W^{-1/2}g.
\end{align*}

\subsection{OLS intercept}

By \eqref{eq:ols_intercept}, the intercept is $\theta_0=\bar Y-\bar X^\top\theta_1$. Averaging \eqref{eq:y_condensed} over the $n$ rows gives
\[
\bar Y
=
\beta_0+\bar X^\top\beta_1+\bar\varepsilon,
\]
where $\bar\varepsilon=\frac{1}{n}\mathbf{1}^\top\varepsilon$ is the mean regression error.
Therefore
\begin{align}
\theta_0
&=
\beta_0+\bar X^\top\beta_1+\bar\varepsilon-\bar X^\top\theta_1
\nonumber\\
&=
\beta_0+\bar\varepsilon-\bar X^\top(\theta_1-\beta_1).
\label{eq:intercept_start}
\end{align}

Besides the slope noise $g$, the regression errors enter the fitted intercept through their mean $\bar\varepsilon$. As the mean of $n$ independent $\mathcal{N}(0,\sigma^2)$ errors, $\bar\varepsilon\sim\mathcal{N}(0,\sigma^2/n)$. We obtain a fourth auxiliary variable, whose distribution, like that of $g$, does not depend on the population-level variables, by dividing $\bar\varepsilon$ by its standard deviation $\sigma/\sqrt n$,
\[
h=\frac{\sqrt n}{\sigma}\bar\varepsilon\sim\mathcal{N}(0,1).
\]
Thus
\begin{equation}
\bar\varepsilon=\frac{\sigma}{\sqrt n}h.
\label{eq:h_definition}
\end{equation}

The last term in \eqref{eq:intercept_start} adjusts the fitted intercept for the error in the fitted slope. To preserve the joint distribution of $(\theta_0,\theta_1)$, this term must use the same auxiliary vector $g$ as the slope.

Using \eqref{eq:xbar_reduced}, \eqref{eq:h_definition}, and $\theta_1-\beta_1=\sigma D_X^{-1}W^{-1/2}g$,
equation \eqref{eq:intercept_start} becomes
\begin{align*}
\theta_0
&=
\beta_0+\frac{\sigma}{\sqrt n}h
-
\sigma
\left(
\vec{\alpha}_{X\mu}+\frac{1}{n}D_Xs
\right)^\top
D_X^{-1}W^{-1/2}g
\\
&=
\beta_0+\frac{\sigma}{\sqrt n}h
-
\sigma
\left(
\vec{\alpha}_{X\mu}^{\top}D_X^{-1}
+\frac{1}{n}s^\top
\right)
W^{-1/2}g.
\end{align*}

It remains to record the independence structure involving $h$. The variable $h$ is a function of $P\varepsilon$, while $g$ is generated from the centered component $M\varepsilon$ together with $Z$. Since $P$ and $M$ are orthogonal projections and $\varepsilon$ is Gaussian, $P\varepsilon$ and $M\varepsilon$ are independent. Moreover, $\varepsilon\perp Z$ implies that the pair $(P\varepsilon,M\varepsilon)$ is independent of $Z$. Consequently, $P\varepsilon$ is independent of $(M\varepsilon,Z)$, and hence
\[
h\perp(g,s,W).
\]
Together with $g\perp(s,W)$ and $s\perp W$, the four auxiliary variables are mutually independent.

\subsection{Final reduced representation}

The full plate can be replaced by the following mutually independent auxiliary variables:
\[
s\sim\mathcal{N}_k(0,nI_k),
\qquad
W\sim\operatorname{Wishart}_k(I_k,n-1),
\qquad
g\sim\mathcal{N}_k(0,I_k),
\qquad
h\sim\mathcal{N}(0,1).
\]
Given these variables and the population-level variables, the fitted ML model parameters are
\begin{align}
\theta_1
&=
\beta_1+\sigma D_X^{-1}W^{-1/2}g,
\label{eq:final_theta1}\\
\theta_0
&=
\beta_0+\frac{\sigma}{\sqrt n}h
-
\sigma\left(
\vec\alpha_{X\mu}^{\top}D_X^{-1}
+\frac{1}{n}s^\top
\right)W^{-1/2}g,
\label{eq:final_theta0}
\end{align}
where
\[
\beta_1=N_b+N_EN_d,
\qquad
\beta_0=\alpha_{Y\mu}+N_E\alpha_{A\mu},
\qquad
\sigma^2=N_E^2\alpha_{A\sigma}^2+\alpha_{Y\sigma}^2.
\]

Equations \eqref{eq:final_theta1}--\eqref{eq:final_theta0} define the same conditional distribution for the OLS coefficients as the original $n$-row plate. Therefore the likelihood of
$\theta=(\theta_0,\theta_1)$ in $\mathcal{A}_{Hist}$ can be computed from the population-level variables and the auxiliary tuple $(s,W,g,h)$ instead of the full plate.

The number of scalar auxiliary variables is
\[
\underbrace{k}_{s}
+
\underbrace{\frac{k(k+1)}{2}}_{W}
+
\underbrace{k}_{g}
+
\underbrace{1}_{h}
=
\frac{k^2+5k+2}{2}.
\]
This proves the lemma.
\end{proof}

\subsection{Second-stage extension: single-prediction collapsed likelihood}
\label{app:single_prediction_collapsed}

Lemma~\ref{thm:suff_stats_app} preserves the full fitted linear model $(\theta_0,\theta_1)$. In the special case where the agent observes only one scalar prediction
\[
Pred^o=\theta_0+x^\top\theta_1
\]
for a fixed observed covariate vector $X^o=x$, a further likelihood-level collapse is possible. This second-stage collapse preserves the scalar likelihood of the observed prediction rather than the full fitted coefficient vector. The collapsed likelihood is obtained by substituting the reduced representation \eqref{eq:final_theta1}--\eqref{eq:final_theta0} into $Pred^o$.

Assume, for this subsection only, that the coefficient priors factorize identically across covariates:
\[
(N_{b,j},N_{d,j})
\overset{iid}{\sim}
\mathcal{N}_2
\left(
\begin{bmatrix}
m_b\\
m_d
\end{bmatrix},
\begin{bmatrix}
v_b & v_{bd}\\
v_{bd} & v_d
\end{bmatrix}
\right),
\qquad
j=1,\ldots,k,
\]
independently of the plate noises and of the remaining population-level variables. This i.i.d.\ form is assumed only so that the marginalized coefficient prior enters through two scalar functions of $x$. The reduction itself does not depend on it (Appendix~\ref{app:single_prediction_general}, which also covers fixed $N_b$ and $N_d$). Define
\[
S^o=\mathbf{1}^\top x,
\qquad
Q^o=x^\top x,
\qquad
\Delta_X^2=(x-\vec{\alpha}_{X\mu})^\top D_X^{-2}(x-\vec{\alpha}_{X\mu}).
\]
We write $\Delta_X=\sqrt{\Delta_X^2}\geq0$ and use $\chi_k'^2(\lambda)$ to denote the non-central chi-squared distribution with $k$ degrees of freedom and non-centrality $\lambda$. Here $S^o$ and $Q^o$ will carry the marginalized coefficient prior, while $\Delta_X^2$ is the squared standardized (Mahalanobis) distance of the query point $x$ from the covariate mean $\vec{\alpha}_{X\mu}$, a measure of how far the prediction is extrapolated. Below, $n\Delta_X^2$ is the non-centrality parameter used to represent $R$. The symbol $\mathbf{1}$ in $S^o=\mathbf{1}^\top x$ denotes the all-ones vector in covariate space $\mathbb{R}^k$. The superscript $o$ marks quantities evaluated at the new patient's covariate vector $x$, the observed value of $X^o$.

\begin{theorem}[Single-prediction collapsed likelihood]
\label{thm:single_prediction_collapsed}
Under the assumptions of Lemma~\ref{thm:suff_stats_app} and the i.i.d.\ Gaussian coefficient prior above, consider a single observed prediction $Pred^o$ for a new patient who is independent of the training data given the population-level variables. Let
\[
R\sim\frac{1}{n}\chi_k'^2(n\Delta_X^2),
\qquad
T\sim\chi^2_{n-k},
\qquad
R\perp T.
\]
Then, after marginalizing $N_b$ and $N_d$, the likelihood of $Pred^o$ under the full plate can be written as
\[
p\left(Pred^o\mid N_E,X^o=x,\text{ remaining population-level variables}\right)
=
\mathbb{E}_{R,T}\left[\mathcal{N}\left(Pred^o;\mu_{\mathrm{red}},V_{\mathrm{red}}\right)\right],
\]
where
\[
\mu_{\mathrm{red}}
=
\beta_0+(m_b+N_E m_d)S^o
\]
and
\[
V_{\mathrm{red}}
=
Q^o(v_b+2N_Ev_{bd}+N_E^2v_d)
+
\sigma^2\left(\frac{1}{n}+\frac{R}{T}\right).
\]
The Gaussian density inside the expectation is the conditional density of $Pred^o$ given $R$ and $T$,
\[
\boxed{Pred^o\mid R,T,N_E,X^o=x,\text{ remaining population-level variables}}
\sim
\mathcal{N}(\mu_{\mathrm{red}},V_{\mathrm{red}}).
\]
The variable $R$ can also be generated from central scalar primitives as
\[
R=
\left(\Delta_X+\frac{G_R}{\sqrt n}\right)^2+\frac{C_R}{n},
\qquad
G_R\sim\mathcal{N}(0,1),
\qquad
C_R\sim\chi^2_{k-1},
\]
where $G_R$, $C_R$ and $T$ are mutually independent, with the convention $\chi^2_0=0$ when $k=1$.
\end{theorem}

\begin{proof}
Condition on $N_E$, the remaining population-level variables, and $X^o=x$. Keep
\[
\beta_1=N_b+N_EN_d
\]
fixed until the last step, where $N_b$ and $N_d$ are integrated out. This lets us apply Lemma~\ref{thm:suff_stats_app}, which holds conditionally on the population coefficients.

From Lemma~\ref{thm:suff_stats_app},
\[
\theta_1
=
\beta_1+\sigma D_X^{-1}W^{-1/2}g
\]
and
\[
\theta_0
=
\beta_0+\frac{\sigma}{\sqrt n}h
-
\sigma\left(
\vec{\alpha}_{X\mu}^{\top}D_X^{-1}
+\frac{1}{n}s^\top
\right)W^{-1/2}g.
\]
Substituting into $Pred^o=\theta_0+x^\top\theta_1$ gives
\begin{align}
Pred^o
&=
\beta_0+x^\top\beta_1
+
\sigma
\left[
\frac{1}{\sqrt n}h
+
\left(
x^\top D_X^{-1}
-
\vec{\alpha}_{X\mu}^{\top}D_X^{-1}
-
\frac{1}{n}s^\top
\right)
W^{-1/2}g
\right].
\end{align}
In the coefficient of $W^{-1/2}g$, the first two terms combine into the standardized deviation of $x$ from the covariate mean, the vector whose squared norm is $\Delta_X^2$,
\[
\delta=D_X^{-1}(x-\vec{\alpha}_{X\mu}).
\]
Since $D_X^{-1}$ is diagonal, hence symmetric, $x^\top D_X^{-1}-\vec{\alpha}_{X\mu}^{\top}D_X^{-1}=\delta^\top$. The whole coefficient of $W^{-1/2}g$ is then $r^\top$, with
\[
r=\delta-\frac{1}{n}s.
\]
By \eqref{eq:xbar_reduced}, $r=D_X^{-1}(x-\bar X)$ is the standardized deviation of $x$ from the sample mean.
Then
\begin{equation}
Pred^o
=
\beta_0+x^\top\beta_1
+
\sigma
\left(
\frac{1}{\sqrt n}h+r^\top W^{-1/2}g
\right).
\label{eq:single_pred_noise_start}
\end{equation}

Conditionally on $r$ and $W$, the term $r^\top W^{-1/2}g=(W^{-1/2}r)^\top g$ is a linear functional of $g\sim\mathcal{N}_k(0,I_k)$, hence Gaussian with zero mean and conditional variance
\begin{align*}
\operatorname{Var}\big(r^\top W^{-1/2}g\mid r,W\big)
&=(W^{-1/2}r)^\top(W^{-1/2}r)\\
&=r^\top W^{-1}r.
\end{align*}
Because $h$ is independent of $(g,r,W)$, the variances of the two zero-mean Gaussian terms add:
\begin{equation}
\frac{1}{\sqrt n}h+r^\top W^{-1/2}g
\mid r,W
\sim
\mathcal{N}
\left(
0,
\frac{1}{n}+r^\top W^{-1}r
\right).
\label{eq:single_pred_noise_conditional}
\end{equation}

By \eqref{eq:single_pred_noise_conditional}, given $(r,W)$ the prediction noise is Gaussian, and the training data enter its variance only through $r^\top W^{-1}r$. We now show that $r^\top W^{-1}r\overset{d}{=}R/T$, with $R=\|r\|^2$ and an independent $T\sim\chi^2_{n-k}$.

We first derive the distribution of $R=\|r\|^2$. Since
$s\sim\mathcal{N}_k(0,nI_k)$,
\[
r=\delta-\frac{1}{n}s\sim\mathcal{N}_k\left(\delta,\frac{1}{n}I_k\right).
\]
The non-central $\chi^2$ distribution describes the squared norm of a Gaussian vector with identity covariance, so we rescale $r$ by $\sqrt n$,
\[
\sqrt n\,r\sim\mathcal{N}_k(\sqrt n\,\delta,I_k).
\]
Since $\|\sqrt n\,r\|^2=nR$, the variable $nR$ has a non-central $\chi^2_k$ distribution with non-centrality $\|\sqrt n\,\delta\|^2=n\Delta_X^2$. Dividing by $n$,
\[
R\sim\frac{1}{n}\chi_k'^2(n\Delta_X^2).
\]

A non-central $\chi^2$ generator is not always available, so we also give an equivalent construction from central primitives. It follows by rotation, and its $k=1$ case is used in the implementation described below. Let
\[
\eta=\sqrt n(r-\delta)\sim\mathcal{N}_k(0,I_k).
\]
Then
\[
nR
=
\|\sqrt n\,\delta+\eta\|^2.
\]
Choose an orthogonal matrix $O_\delta$ that maps $\delta$ to $\Delta_X e_1$, where $e_1=(1,0,\ldots,0)^\top\in\mathbb R^k$. If $\delta=0$, any orthogonal $O_\delta$ may be used. By rotational invariance,
\[
O_\delta\eta\sim\mathcal{N}_k(0,I_k).
\]
Write
\[
O_\delta\eta=(G_R,C_2,\ldots,C_k)^\top,
\]
with independent standard normal components. Then
\[
nR
=
(\sqrt n\,\Delta_X+G_R)^2+\sum_{j=2}^kC_j^2.
\]
Thus
\[
R=
\left(\Delta_X+\frac{G_R}{\sqrt n}\right)^2+\frac{C_R}{n},
\qquad
C_R=\sum_{j=2}^kC_j^2\sim\chi^2_{k-1}.
\]

We next reduce $r^\top W^{-1}r$. Since $W\perp s$, also $W\perp r$, so conditioning on $r$ leaves the distribution of $W$ unchanged. Conditionally on $r$, if $r=0$ the quadratic form is zero. If $r\neq0$, choose an orthogonal matrix $O_r$ such that
\[
O_r^\top r=\|r\|e_1.
\]
Since the Wishart distribution with identity scale is invariant under orthogonal conjugation,
\[
O_r^\top WO_r\sim\operatorname{Wishart}_k(I_k,n-1).
\]
Therefore, conditionally on $r$,
\[
r^\top W^{-1}r
=
\|r\|^2 e_1^\top (O_r^\top WO_r)^{-1}e_1
\overset{d}{=}
\|r\|^2 e_1^\top W^{-1}e_1.
\]

The distribution of the reciprocal $[(W^{-1})_{11}]^{-1}$ of the diagonal entry of an identity-scale inverse Wishart is classical \citep[Theorem~3.2.12]{muirhead1982}; for completeness we give a short self-contained derivation. Let $\nu=n-1$ and represent $W=V^\top V$, where $V\in\mathbb{R}^{\nu\times k}$ has independent standard normal entries. Partition
\[
V=[v_1,V_2],
\]
where $v_1\in\mathbb{R}^{\nu}$ and $V_2\in\mathbb{R}^{\nu\times(k-1)}$. The Schur complement identity gives
\[
\left[(W^{-1})_{11}\right]^{-1}
=
v_1^\top
\left[
I_\nu-V_2(V_2^\top V_2)^{-1}V_2^\top
\right]
v_1.
\]
For $k=1$, $V_2$ has no columns, and the projection onto its column span is zero, so the bracketed matrix is $I_\nu$. In general, the bracketed matrix is the orthogonal projector onto the complement of the column span of $V_2$. Almost surely, $V_2$ has rank $k-1$, so this projector has rank
\[
\nu-(k-1)=\nu-k+1.
\]
Conditionally on $V_2$, the projector is fixed and $v_1\sim\mathcal{N}_\nu(0,I_\nu)$. In an orthonormal basis adapted to this projector, the quadratic form is the sum of squares of $\nu-k+1$ independent standard normal coordinates. Hence
\[
\left[(W^{-1})_{11}\right]^{-1}
\sim
\chi^2_{\nu-k+1}.
\]
Substituting $\nu=n-1$ gives
\[
\nu-k+1=n-k.
\]
Therefore
\[
e_1^\top W^{-1}e_1
\overset{d}{=}
\frac{1}{T},
\qquad
T\sim\chi^2_{n-k}.
\]
Because $r\sim\mathcal{N}_k(\delta,\frac{1}{n}I_k)$, we have $r\neq0$ almost surely, and the ratio
\[
T=\frac{\|r\|^2}{r^\top W^{-1}r}
\]
is well defined almost surely. Conditionally on $r$, $T\sim\chi^2_{n-k}$, and this distribution does not depend on $r$. Hence $T$ is independent of $r$, and therefore of $R=\|r\|^2$. Thus
\[
r^\top W^{-1}r
\overset{d}{=}
\frac{R}{T},
\qquad
T\sim\chi^2_{n-k},
\qquad
T\perp R.
\]

By \eqref{eq:single_pred_noise_start}--\eqref{eq:single_pred_noise_conditional}, the conditional prediction mean is $\beta_0+x^\top\beta_1$, and its variance depends on $(r,W)$ only through $r^\top W^{-1}r=R/T$. Averaging over $(r,W)$ at fixed $(R,T)$ therefore leaves the same Gaussian distribution. Conditional also on $\beta_1$, $N_E$, $X^o=x$ and the remaining population-level variables,
\[
Pred^o\mid\beta_1,R,T
\sim\mathcal N\!\left(\beta_0+x^\top\beta_1,\,
\sigma^2\left(\frac1n+\frac RT\right)\right).
\]

It remains to integrate out $N_b$ and $N_d$. In the conditional distribution above the coefficients enter only through the mean $\beta_0+x^\top\beta_1$, so marginalizing their prior adds an independent Gaussian term to the prediction and leaves the OLS-noise variance $\sigma^2(\tfrac1n+R/T)$ unchanged. Define
\[
B^o=x^\top N_b,
\qquad
D^o=x^\top N_d.
\]
Then
\[
x^\top\beta_1
=
B^o+N_ED^o.
\]
Under the i.i.d.\ Gaussian coefficient prior,
\[
(B^o,D^o)
\]
is bivariate Gaussian with
\[
\mathbb{E}[B^o]=m_bS^o,
\qquad
\mathbb{E}[D^o]=m_dS^o,
\]
\[
\operatorname{Var}(B^o)=v_bQ^o,
\qquad
\operatorname{Var}(D^o)=v_dQ^o,
\qquad
\operatorname{Cov}(B^o,D^o)=v_{bd}Q^o.
\]
Therefore, conditionally on $N_E$,
\[
B^o+N_ED^o
\sim
\mathcal{N}
\left(
(m_b+N_Em_d)S^o,
Q^o(v_b+2N_Ev_{bd}+N_E^2v_d)
\right).
\]
This Gaussian term is independent of the OLS noise represented by $(R,T)$, because the coefficient priors are independent of the auxiliary variables replacing the plate. Summing the two independent Gaussian components yields
\[
Pred^o
\mid
R,T,N_E,X^o=x,\text{ remaining population-level variables}
\sim
\mathcal{N}(\mu_{\mathrm{red}},V_{\mathrm{red}}),
\]
with
\[
\mu_{\mathrm{red}}
=
\beta_0+(m_b+N_E m_d)S^o
\]
and
\[
V_{\mathrm{red}}
=
Q^o(v_b+2N_Ev_{bd}+N_E^2v_d)
+
\sigma^2\left(\frac{1}{n}+\frac{R}{T}\right).
\]
Each step above preserves the distribution of $Pred^o$ given $N_E$, $X^o=x$ and the remaining population-level variables. Integrating over the distributions of $R$ and $T$ therefore gives the likelihood of $Pred^o$ under the full plate,
\[
p\left(Pred^o\mid N_E,X^o=x,\text{ remaining population-level variables}\right)
=
\mathbb{E}_{R,T}\left[\mathcal{N}\left(Pred^o;\mu_{\mathrm{red}},V_{\mathrm{red}}\right)\right].
\]
This proves the single-prediction collapsed likelihood.
\end{proof}

The theorem above gives the conditional likelihood of the observed scalar prediction given the observed covariate vector. The full Bayesian update for the agent also contains the likelihood of the observed covariate vector itself. Under the diagonal covariate model,
\[
p(X^o=x\mid \vec\alpha_{X\mu},D_X)
=
\prod_{j=1}^k
\mathcal{N}\!\left(x_j;\alpha_{X\mu,j},\alpha_{X\sigma,j}^2\right).
\]
This term updates the agent's beliefs about the covariate population parameters. It is not absorbed into the collapsed likelihood for $Pred^o$.

The collapse also retains the information needed for the treatment-effect decision rule. For a fixed realization $\psi$ of the population-level parameters, the interventional model used by the agent gives
\[
\mathbb{E}^{\mathcal{A}_{Inf}^\psi}[Y\mid X=x,do(A=a)]
=
B^o+N_Ea+\alpha_{Y\mu},
\]
where $B^o=x^\top N_b$. Therefore, for two treatment levels $a_1$ and $a_2$,
\[
\text{CATE}^{\psi}
=
\mathbb{E}^{\mathcal{A}_{Inf}^\psi}[Y\mid X=x,do(A=a_1)]
-
\mathbb{E}^{\mathcal{A}_{Inf}^\psi}[Y\mid X=x,do(A=a_2)]
=
N_E(a_1-a_2).
\]
This is the CATE at fixed parameters. The agent's estimate is its posterior mean, $(a_1-a_2)\mathbb E[N_E\mid X^o=x,Pred^o]$. The projection $B^o$ cancels in the contrast. Hence the reduced likelihood must preserve the posterior information about $N_E$ for the decision rule, but it does not need to preserve the full posterior over the individual coefficient vectors $N_b$ and $N_d$.

\subsection{General prior across covariates}
\label{app:single_prediction_general}

The compact form of Theorem~\ref{thm:single_prediction_collapsed} uses the i.i.d.\ coefficient prior, but the reduction itself does not require it. With a jointly Gaussian prior on $(N_b,N_d)$, independent of the other population-level variables and of the standardized plate noises, having mean vectors $m_b,m_d\in\mathbb R^k$ and covariance blocks $\Sigma_b,\Sigma_d,\Sigma_{bd}\in\mathbb R^{k\times k}$,
\[
N_b\sim\mathcal{N}_k({m}_b,\Sigma_b),
\qquad
N_d\sim\mathcal{N}_k({m}_d,\Sigma_d),
\qquad
\operatorname{Cov}(N_b,N_d)=\Sigma_{bd},
\]
the projections $B^o=x^\top N_b$ and $D^o=x^\top N_d$ used in the last step of the proof are still jointly Gaussian, with
\[
\mathbb{E}[B^o]=x^\top{m}_b,
\qquad
\mathbb{E}[D^o]=x^\top{m}_d,
\]
\[
\operatorname{Var}(B^o)=x^\top\Sigma_b\,x,
\qquad
\operatorname{Var}(D^o)=x^\top\Sigma_d\,x,
\qquad
\operatorname{Cov}(B^o,D^o)=x^\top\Sigma_{bd}\,x.
\]
The shorthands $S^o=\mathbf{1}^\top x$ and $Q^o=x^\top x$ of the theorem are replaced by these linear and quadratic forms in $x$, but the structure of the result is identical. More generally, the same Gaussian formulas apply if $(N_b,N_d)$ is jointly Gaussian conditional on the remaining population-level variables, with the corresponding conditional moments, and remains conditionally independent of the standardized plate noises and the new patient. In particular, if the prior depends only on $N_E$, use its moments conditional on $N_E$.

For fixed coefficients $N_b=b$ and $N_d=d$, as in our experiments, set ${m}_b=b$, ${m}_d=d$ and $\Sigma_b=\Sigma_d=\Sigma_{bd}=0$. Then $B^o$ and $D^o$ are constants, and the conditional Gaussian prediction likelihood has mean and variance
\begin{align*}
\mu_{\mathrm{red}}&=\beta_0+x^\top(b+N_Ed),\\
V_{\mathrm{red}}&=\sigma^2\left(\frac1n+\frac RT\right).
\end{align*}

For non-Gaussian coefficient priors, the reduction to $(R,T)$ still holds under the same conditional independence assumptions. At fixed $N_E$, remaining population-level variables and $X^o=x$, the likelihood conditional on $(R,T)$ is the Gaussian likelihood from the proof averaged over the conditional distribution of $B^o+N_ED^o$. This average need not have a closed form.

\subsection{Variable count and implementation notes}
\label{app:single_prediction_count}

Table~\ref{tab:single_prediction_count} summarizes how many scalar sampling events are needed to represent the training-data plate under the three representations derived above. The full plate samples the $n$ historical units explicitly, with $k+2$ scalars per unit. The reduced plate of Lemma~\ref{thm:suff_stats_app} is constant in $n$ but quadratic in $k$ through the Wishart matrix $W$. The likelihood-level auxiliaries $(R,T)$ of Theorem~\ref{thm:single_prediction_collapsed} are constant in both. Population-level variables outside the plate are excluded from the count, since they are common to the three representations.

\begin{table}[h]
\centering
\begin{tabular}{cccc}
\toprule
$k$ & Full plate ($n=1000$) & Reduced plate (Lemma~\ref{thm:suff_stats_app}) & Collapsed (Thm~\ref{thm:single_prediction_collapsed}) \\
\midrule
1  & 3000  & 4    & 2 \\
2  & 4000  & 8    & 2 \\
5  & 7000  & 26   & 2 \\
10 & 12000 & 76   & 2 \\
\bottomrule
\end{tabular}
\caption{Number of scalar sampling events needed to represent the training-data plate, as a function of the number of covariates $k$, for the three plate representations.}
\label{tab:single_prediction_count}
\end{table}

Probabilistic programming languages do not always provide a native non-central chi-squared distribution (PyMC, for instance, does not). At $k=1$, our implementation samples $G_R\sim\mathcal N(0,1)$ and sets $R=(\delta+G_R/\sqrt n)^2$, where $\delta=(x-\alpha_{X\mu})/\alpha_{X\sigma}$ is the signed standardized offset. By the symmetry of $G_R$, this has the same distribution as the construction in Theorem~\ref{thm:single_prediction_collapsed} with $C_R=0$. At $k\geq2$, it samples $R$ from its scaled non-central $\chi^2$ distribution, implemented in our code. In both cases, $T\sim\chi^2_{n-k}$ independently.

In addition, $\alpha_{A\mu}$, $\alpha_{Y\mu}$ and any uncertain coefficients entering the prediction linearly have Gaussian priors independent of one another and of the other population-level variables. These quantities are marginalized analytically conditional on $N_E$ and the remaining sampled variables, while $\vec\alpha_{X\mu}$ remains sampled. The two signs of $N_E$ are then summed out exactly at fixed $|N_E|$ to improve NUTS mixing when the posterior has separated positive and negative modes. After sampling, the sign and marginalized variables are reconstructed from their conditional posterior, preserving the target posterior. The reported experiments fix $N_b$ and $N_d$.

\subsection{No-intercept addendum}
\label{app:no_intercept}

For the no-intercept model, we reuse the standardized design, the auxiliary variables $s$ and $W$, and the condensed outcome representation from the proof of Lemma~\ref{thm:suff_stats_app}. We derive its single-prediction likelihood using $(s,W)$, without the further reduction to $(R,T)$.

\begin{corollary}[No-intercept case]
\label{corollary_no_intercept}
Under the assumptions of Lemma~\ref{thm:suff_stats_app}, except that the ML model is fitted by ordinary least squares without an intercept, the joint distribution of its fitted coefficients conditional on the population-level variables has an exact representation using $\frac{k^2+5k}{2}$ scalar auxiliary variables.
\end{corollary}

\begin{proof}
The SCM, standardization, and condensed outcome representation are the same as in Lemma~\ref{thm:suff_stats_app}. In particular, by
\eqref{eq:design_decomposition} and \eqref{eq:y_condensed},
\begin{align*}
X&=\mathbf{1}\vec{\alpha}_{X\mu}^{\top}+ZD_X,\\
Y&=\beta_0\mathbf{1}+X\beta_1+\varepsilon,
\end{align*}
with $\varepsilon\sim\mathcal{N}_n(0,\sigma^2I_n)$ and $\varepsilon\perp Z$.
The no-intercept estimator, which plays the role of $\theta$ in this model, is
\[
\phi=(X^\top X)^{-1}X^\top Y.
\]

The uncentered covariate sum is
\[
S_x
:=
X^\top\mathbf{1}
=
n\vec{\alpha}_{X\mu}+D_Xs.
\]
Moreover,
\[
Z^\top Z
=
Z^\top PZ+Z^\top MZ
=
\frac{1}{n}ss^\top+W.
\]
Therefore
\begin{align}
X^\top X
&=
\frac{1}{n}S_xS_x^\top+D_XWD_X.
\label{eq:no_intercept_xx}
\end{align}
Since $W$ is positive definite almost surely and $D_X$ is invertible, $D_XWD_X$ is positive definite almost surely. The rank-one matrix $S_xS_x^\top/n$ is positive semidefinite, so $X^\top X$ is positive definite almost surely.

Substituting the condensed outcome equation,
\begin{align}
\phi
&=
(X^\top X)^{-1}X^\top(\beta_0\mathbf{1}+X\beta_1+\varepsilon)
\nonumber\\
&=
\beta_1
+
\beta_0(X^\top X)^{-1}S_x
+
(X^\top X)^{-1}X^\top\varepsilon.
\label{eq:no_intercept_start}
\end{align}
Conditionally on $Z$,
\[
X^\top\varepsilon\mid Z
\sim
\mathcal{N}_k(0,\sigma^2X^\top X).
\]
The covariance is a deterministic function of $(s,W)$ through \eqref{eq:no_intercept_xx}. As in the definition of $g$ in the proof of Lemma~\ref{thm:suff_stats_app}, define
\[
\tilde g
=
\sigma^{-1}(X^\top X)^{-1/2}X^\top\varepsilon,
\qquad\text{so that}\qquad
X^\top\varepsilon
=
\sigma(X^\top X)^{1/2}\tilde g.
\]
Conditionally on $Z$, $\tilde g$ is Gaussian with mean $0$ and covariance $I_k$; since this conditional distribution does not depend on $Z$,
\[
\tilde g\sim\mathcal{N}_k(0,I_k),
\qquad
\tilde g\perp(s,W).
\]
Substituting into \eqref{eq:no_intercept_start},
\[
\phi
=
\beta_1
+
\beta_0(X^\top X)^{-1}S_x
+
\sigma(X^\top X)^{-1/2}\tilde g,
\]
with
\[
S_x=n\vec{\alpha}_{X\mu}+D_Xs,
\qquad
X^\top X=\frac{1}{n}S_xS_x^\top+D_XWD_X.
\]

Thus the no-intercept plate is represented by
\[
s\sim\mathcal{N}_k(0,nI_k),
\qquad
W\sim\operatorname{Wishart}_k(I_k,n-1),
\qquad
\tilde g\sim\mathcal{N}_k(0,I_k),
\]
with $s\perp W$ and $\tilde g\perp(s,W)$. The scalar count is
\[
\underbrace{k}_{s}
+
\underbrace{\frac{k(k+1)}{2}}_{W}
+
\underbrace{k}_{\tilde g}
=
\frac{k^2+5k}{2}.
\]
No separate scalar $h$ is needed here. The mean and centered components of the regression error enter $\phi$ only through $X^\top\varepsilon$ and are represented jointly by the Gaussian vector $\tilde g$.
\end{proof}

\paragraph{Single prediction without intercept.}
Our no-intercept experiment (Appendix~\ref{app:no_intercept_experiments}) observes a single prediction $Pred_{\mathrm{no}}^o=x^\top\phi$ at $X^o=x$, with $x\neq0$. Its likelihood follows from the representation above by marginalizing $\tilde g$. Define
\begin{align*}
\lambda_x&=x^\top(X^\top X)^{-1}S_x,\\
q_x&=x^\top(X^\top X)^{-1}x.
\end{align*}
Conditionally on the population-level variables and on $(s,W)$, the prediction is linear in the standard Gaussian vector $\tilde g$. Hence
\[
Pred_{\mathrm{no}}^o\mid s,W\sim\mathcal{N}\left(x^\top\beta_1+\beta_0\lambda_x,\ \sigma^2q_x\right).
\]
The likelihood is obtained by integrating this Gaussian distribution over $(s,W)$. This $(s,W)$ representation uses $k+\frac{k(k+1)}{2}$ scalar auxiliary variables. In the univariate experiment, $s\sim\mathcal{N}(0,n)$ and $W\sim\chi^2_{n-1}$ are independent, so this representation uses two scalar auxiliaries.

\section{Perfect predictive calibration does not identify treatment effects}
\label{app:calibration_example}

This example separates predictive calibration from the causal information needed for a treatment decision. Consider
\[
X\sim\mathcal N(0,1),\qquad
A=\mu_A+\varepsilon_A,\qquad
Y=X+N_EA+\varepsilon_Y,
\]
where $\varepsilon_A\sim\mathcal N(0,\sigma_A^2)$ and $\varepsilon_Y\sim\mathcal N(0,\sigma_Y^2)$ are independent of each other and of $X$, and the noise variances are positive. The coefficient of $X$ is known and fixed at one. Higher outcomes are preferred. Here $\mu_A$ denotes the historical treatment mean; the corresponding quantity in the agent's model is $\alpha_{A\mu}$.

\paragraph{The same calibrated predictor, opposite effects.}
The two SCMs $(\mu_A,N_E)=(1,1)$ and $(\mu_A,N_E)=(-1,-1)$ induce the same observational distribution of $(X,Y)$, since in both cases
\[
Y\mid X=x\sim\mathcal N(x+1,\sigma_A^2+\sigma_Y^2).
\]
The observational intercept is the product $N_E\mu_A$, which equals one in both SCMs, while the slope remains one. The noise variance $N_E^2\sigma_A^2+\sigma_Y^2$ is also unchanged.
Their population predictor is $f(x)=x+1$, which is perfectly calibrated in mean:
\[
\mathbb E[Y\mid f(X)]
=\mathbb E[\mathbb E[Y\mid X]\mid f(X)]=f(X).
\]
Nevertheless, for $a_1=1$ and $a_2=0$, the CATE is $N_E$, hence $+1$ in the first SCM and $-1$ in the second. No property of the treatment-naive observational distribution $P(X,Y)$ alone distinguishes these effects. In particular, stronger predictive calibration does not resolve the ambiguity without information about the causal process.

\paragraph{A harmful Bayesian update.}
Take the first SCM to be true, and let the agent observe $X^o=1$ and $Pred^o=f(1)=2$. The agent's treatment-effect prior is $N_E\sim\mathcal N(1,1)$, and it chooses $a_1$ when its posterior mean CATE exceeds $1/2$. Before seeing the prediction its mean CATE is one, so it chooses $a_1$.

For algebraic clarity, take the zero-variance limit of the agent's Gaussian prior on $\alpha_{A\mu}$, concentrated at the incorrect value $-1$, with the other structural quantities known. In the population-predictor limit of OLS training, the agent's model gives $Pred^o=X^o+N_E\alpha_{A\mu}=1-N_E$. Observing $Pred^o=2$ therefore concentrates its posterior at $N_E=-1$ and changes the decision to $a_2$. Under the true SCM,
\[
\mathbb E[Y\mid X=1,do(A=1)]=2,\qquad
\mathbb E[Y\mid X=1,do(A=0)]=1.
\]
The update lowers the expected outcome by one despite perfect predictive calibration. This exact calculation concerns the population predictor and the limiting prior; it does not assert that every finite-sample OLS fit is perfectly calibrated.

\paragraph{What disclosure changes.}
If the historical treatment mean is disclosed and the agent instead fixes $\alpha_{A\mu}=1$, the same observation obeys $Pred^o=1+N_E$, giving $N_E=1$ and the beneficial action $a_1$. The framework thus not only diagnoses the problem but also points to which prior should be corrected to steer the agent toward the beneficial action.

\newpage
\section*{NeurIPS Paper Checklist}

\begin{enumerate}

\item {\bf Claims}
    \item[] Question: Do the main claims made in the abstract and introduction accurately reflect the paper's contributions and scope?
    \item[] Answer: \answerYes{} %
    \item[] Justification: The four contributions of the paper, the framework, the tractability theorem and the two main experimental findings, are all accurately introduced in the Abstract and Introduction. The framework is described in Section \ref{sec:framework}, the theorem is outlined in Subsection \ref{sec:theorem} and described in full detail in Appendix \ref{app:proof_thm_1}. The Experimental setup and results are described in Sections \ref{sec:exp_setup} and \ref{sec:results}. Further details of experimental design are provided in Appendix \ref{app:exp_setup}. The limitations of the contributions are discussed in the Discussion section, clarifying the scope of the claims.

\item {\bf Limitations}
    \item[] Question: Does the paper discuss the limitations of the work performed by the authors?
    \item[] Answer: \answerYes{} %
    \item[] Justification: The limitations of the work are discussed in a dedicated part of the Discussion section. The assumptions of the theorem are stated both in the main text (informally) and detailed in the corresponding section of the Appendix. The scope of the experiments mirrors the scope of the Theorem, as mentioned explicitly in the text.

\item {\bf Theory assumptions and proofs}
    \item[] Question: For each theoretical result, does the paper provide the full set of assumptions and a complete (and correct) proof?
    \item[] Answer: \answerYes{} %
    \item[] Justification: The full description of the assumptions and the proof of Theorem \ref{thm:suff_stats} are given in Appendix \ref{app:proof_thm_1}.

    \item {\bf Experimental result reproducibility}
    \item[] Question: Does the paper fully disclose all the information needed to reproduce the main experimental results of the paper to the extent that it affects the main claims and/or conclusions of the paper (regardless of whether the code and data are provided or not)?
    \item[] Answer: \answerYes{} %
    \item[] Justification: The experiments are described in Section \ref{sec:exp_setup} and all details about the design and implementation that did not fit in the main body are reported in Appendix \ref{app:exp_setup}. Further, we provided access to the code used to run the experiments.

\item {\bf Open access to data and code}
    \item[] Question: Does the paper provide open access to the data and code, with sufficient instructions to faithfully reproduce the main experimental results, as described in supplemental material?
    \item[] Answer: \answerYes{} %
    \item[] Justification: The code used to run the experiments is provided in the submission (\url{https://github.com/thelogicalgrammar/CapturingUsersPredictionModels}). The paper provides a detailed account of the design and implementation.

\item {\bf Experimental setting/details}
    \item[] Question: Does the paper specify all the training and test details (e.g., data splits, hyperparameters, how they were chosen, type of optimizer) necessary to understand the results?
    \item[] Answer: \answerYes{} %
    \item[] Justification: All training and test details are provided in the paper in Section \ref{sec:exp_setup} and Appendix \ref{app:exp_setup}.

\item {\bf Experiment statistical significance}
    \item[] Question: Does the paper report error bars suitably and correctly defined or other appropriate information about the statistical significance of the experiments?
    \item[] Answer: \answerYes{} %
    \item[] Justification: Fig.~\ref{fig:misaligned_ANX_with_intercept}, Fig.~\ref{fig:no_intercept_mu_X} and the appendix figures summarize CATE distributions by means and bands of plus or minus one standard deviation, and downstream outcomes by boxplots. The CATE bands combine within-replicate uncertainty with variation between replicate means. They are not standard errors of the aggregate mean. 

\item {\bf Experiments compute resources}
    \item[] Question: For each experiment, does the paper provide sufficient information on the computer resources (type of compute workers, memory, time of execution) needed to reproduce the experiments?
    \item[] Answer: \answerYes{} %
    \item[] Justification: Information is provided in Section \ref{sec:implementation_details} in the Appendix.
    
\item {\bf Code of ethics}
    \item[] Question: Does the research conducted in the paper conform, in every respect, with the NeurIPS Code of Ethics \url{https://neurips.cc/public/EthicsGuidelines}?
    \item[] Answer: \answerYes{} %
    \item[] Justification: The authors reviewed the Code and confirm the paper complies.

\item {\bf Broader impacts}
    \item[] Question: Does the paper discuss both potential positive societal impacts and negative societal impacts of the work performed?
    \item[] Answer: \answerYes{} %
    \item[] Justification: Despite its formal nature, social impact is a core theme of the paper. The paper captures a previously underexplored learning mechanism for agents interacting with ML-driven decision support, and shows that--when specific misalignments of agent beliefs happen--this type of tool can lead to harmful consequences even in very ideal conditions (simple ML model, simple causal mechanism, fully rational agent, etc.). The paper clearly states that this failure mode applies to all treatment-naive prognostic decision support tools that fit our definition, in all fields of application, including high-stakes ones such as healthcare or law enforcement. Furthermore, the paper stresses the need to counter this failure mode by properly training the users of such decision support tools.
    The potential positive impact of the paper is a heightened attention to the themes above, as well as suggestions on how to conduct pre-deployment audits.
    
\item {\bf Safeguards}
    \item[] Question: Does the paper describe safeguards that have been put in place for responsible release of data or models that have a high risk for misuse (e.g., pre-trained language models, image generators, or scraped datasets)?
    \item[] Answer: \answerNA{} %
    \item[] Justification: The paper is not releasing a new model or data.

\item {\bf Licenses for existing assets}
    \item[] Question: Are the creators or original owners of assets (e.g., code, data, models), used in the paper, properly credited and are the license and terms of use explicitly mentioned and properly respected?
    \item[] Answer: \answerYes{} %
    \item[] Justification: The key packages used are properly referenced in the paper (e.g. PyMC), and the code is released under the MIT license.

\item {\bf New assets}
    \item[] Question: Are new assets introduced in the paper well documented and is the documentation provided alongside the assets?
    \item[] Answer: \answerYes{} %
    \item[] Justification: The code used to generate the experiments is released alongside the paper, and it is properly documented.

\item {\bf Crowdsourcing and research with human subjects}
    \item[] Question: For crowdsourcing experiments and research with human subjects, does the paper include the full text of instructions given to participants and screenshots, if applicable, as well as details about compensation (if any)? 
    \item[] Answer: \answerNA{} %
    \item[] Justification: The paper does not involve crowdsourcing nor research with human subjects.

\item {\bf Institutional review board (IRB) approvals or equivalent for research with human subjects}
    \item[] Question: Does the paper describe potential risks incurred by study participants, whether such risks were disclosed to the subjects, and whether Institutional Review Board (IRB) approvals (or an equivalent approval/review based on the requirements of your country or institution) were obtained?
    \item[] Answer: \answerNA{} %
    \item[] Justification: The paper does not involve crowdsourcing nor research with human subjects.

\item {\bf Declaration of LLM usage}
    \item[] Question: Does the paper describe the usage of LLMs if it is an important, original, or non-standard component of the core methods in this research? Note that if the LLM is used only for writing, editing, or formatting purposes and does \emph{not} impact the core methodology, scientific rigor, or originality of the research, declaration is not required.
    \item[] Answer: \answerNA{} %
    \item[] Justification: LLMs were only used for editing and formatting.

\end{enumerate}
\end{document}